\documentclass[10pt,twocolumn,letterpaper]{article}
\usepackage{cvpr}
\usepackage{graphicx}
\usepackage{amsmath}
\usepackage{amssymb}
\usepackage{booktabs}

\usepackage{multirow}
\usepackage{threeparttable}
\usepackage{textcomp}

\usepackage[pagebackref,breaklinks,colorlinks]{hyperref}

\usepackage[capitalize]{cleveref}
\crefname{section}{Sec.}{Secs.}
\Crefname{section}{Section}{Sections}
\Crefname{table}{Table}{Tables}
\crefname{table}{Tab.}{Tabs.}
\def\cvprPaperID{XXXX} 
\def\confName{CVPR}
\def\confYear{2022}

\begin{document}

\title{GLPanoDepth: Global-to-Local Panoramic Depth Estimation}


\author{Jiayang Bai \textsuperscript{\rm 1}, Shuichang Lai \textsuperscript{\rm 1}, Haoyu Qin\textsuperscript{\rm 1}, Jie Guo\textsuperscript{\rm 1}\textsuperscript{$\dagger$} and Yanwen Guo\textsuperscript{\rm 1}\textsuperscript{$\dagger$}\\
\textsuperscript{\rm 1}State Key Lab for Novel Software Technology, Nanjing University
}

\maketitle

\begin{abstract}
In this paper, we propose a learning-based method for predicting dense depth values of a scene from a monocular omnidirectional image. An omnidirectional image has a full field-of-view, providing much more complete descriptions of the scene than perspective images. However, fully-convolutional networks that most current solutions rely on fail to capture rich global contexts from the panorama. To address this issue and also the distortion of equirectangular projection in the panorama, we propose \emph{Cubemap Vision Transformers} (CViT), a new transformer-based architecture that can model long-range dependencies and extract distortion-free global features from the panorama. We show that cubemap vision transformers have a global receptive field at every stage and can provide globally coherent predictions for spherical signals. To preserve important local features, we further design a convolution-based branch in our pipeline (dubbed \emph{GLPanoDepth}) and fuse global features from cubemap vision transformers at multiple scales. This global-to-local strategy allows us to fully exploit useful global and local features in the panorama, achieving state-of-the-art performance in panoramic depth estimation.
\end{abstract}

\section{Introduction}
The boom of numerous 3D vision applications (\emph{e.g.}, autonomous driving \cite{8025618,8100178,9197319}, 3D scene reconstruction \cite{https://doi.org/10.1111/cgf.13386,https://doi.org/10.1111/cgf.14021} and human motion analysis \cite{CHEN20131995}) has witnessed the effectiveness of adopting depth as an important modality besides RGB color. To obtain the depth information, one could resort to some depth sensors such time-of-flight (TOF) cameras or could infer it from RGB images. Compared with capturing depth from specialized depth sensors, it is more attractive to infer depth from color images since they are still the only sources in many scenarios. Much effort has been devoted to solve this ill-posed problem for single perspective image (image captured by conventional cameras with a limited field-of-view). Among them, deep learning-based methods have achieved very promising results \cite{DBLP:journals/corr/abs-1906-06113,s20082272,MING202114}.

\begin{figure}[tbp]
  \begin{center}
  \renewcommand\tabcolsep{1.0pt}
  \begin{tabular}{ccc}
  
    \includegraphics[width=0.33\linewidth, trim={0px, 0px, 0px, 0px}, clip]{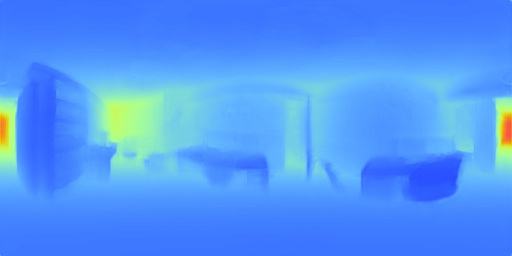}   &
    \includegraphics[width=0.33\linewidth, trim={0px, 0px, 0px, 0px}, clip]{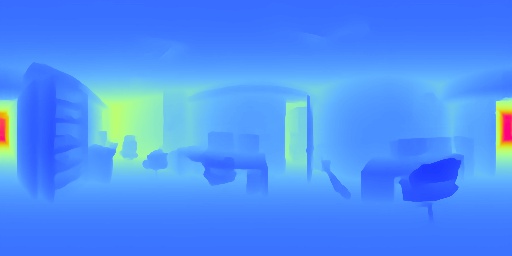}     &
    \includegraphics[width=0.33\linewidth, trim={0px, 0px, 0px, 0px}, clip]{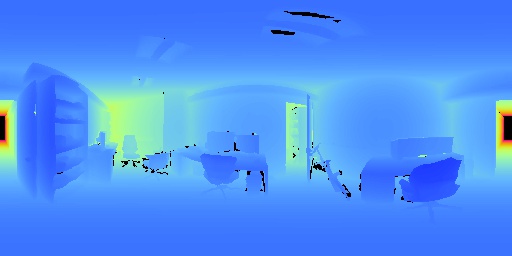}         \\
    \includegraphics[width=0.33\linewidth, trim={0px, 0px, 100px, 0px}, clip]{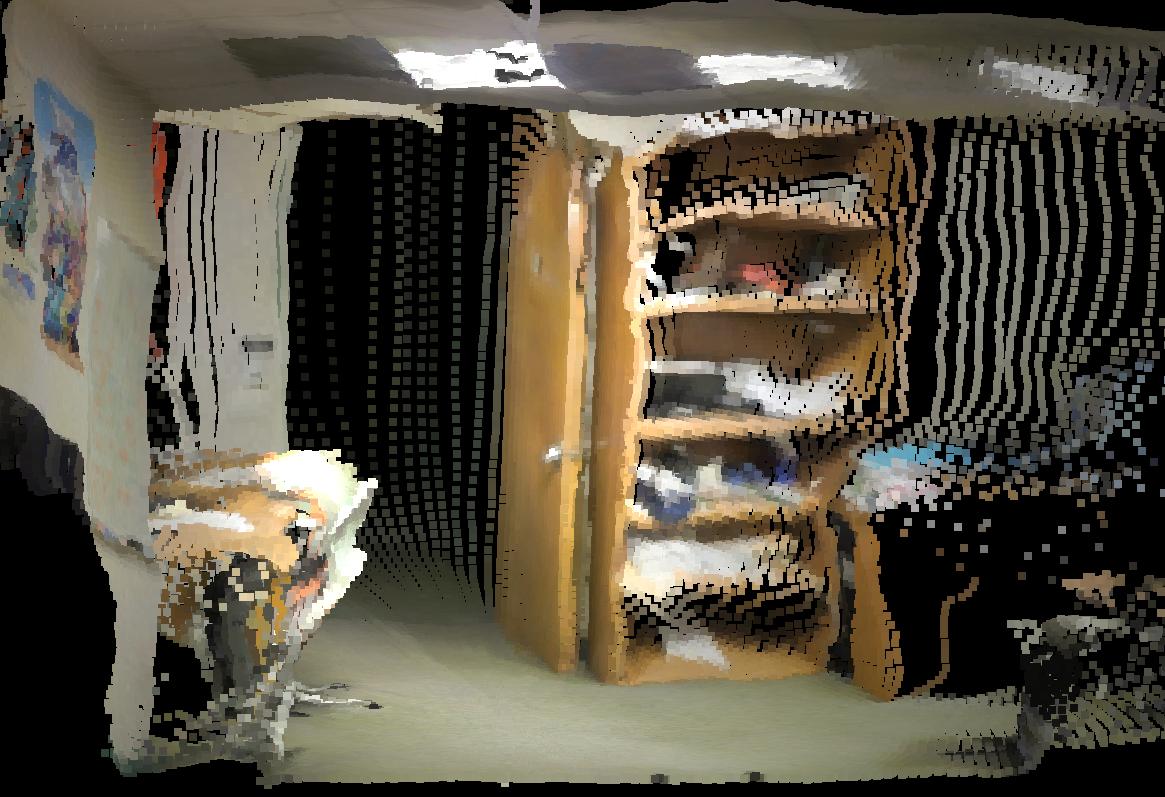}   &
    \includegraphics[width=0.33\linewidth, trim={0px, 0px, 100px, 0px}, clip]{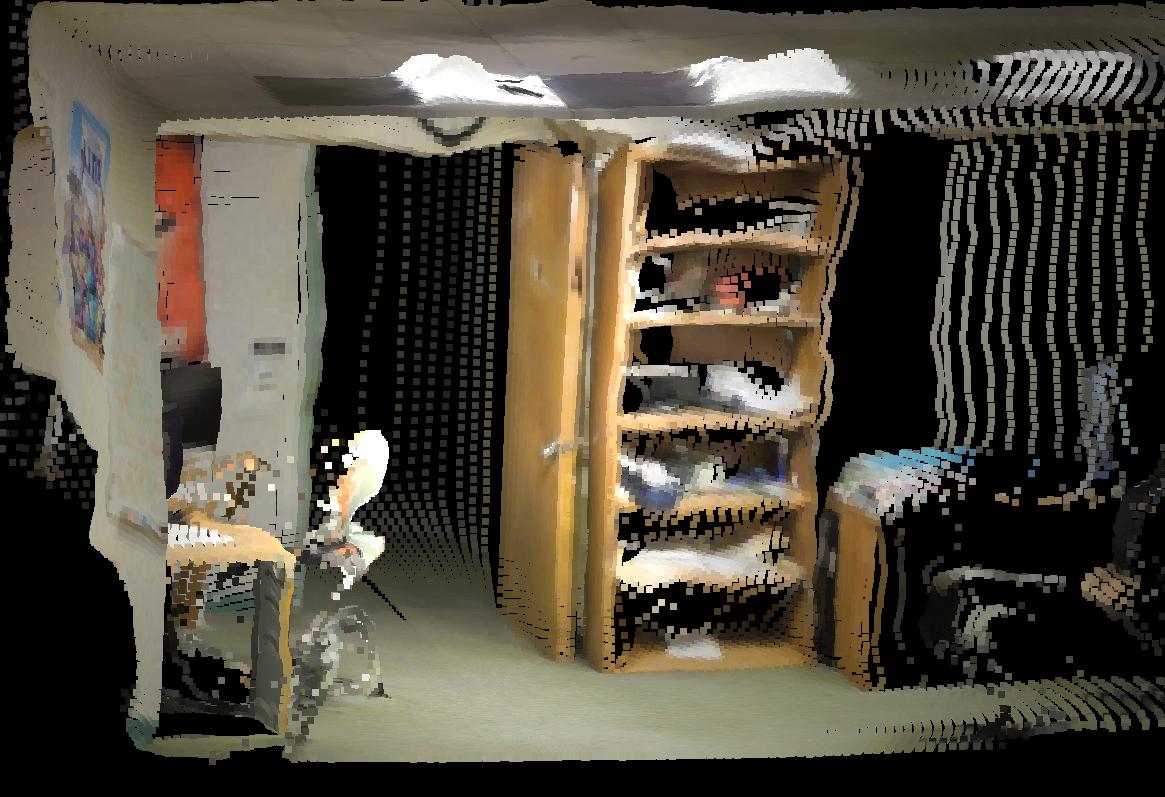}     &
    \includegraphics[width=0.33\linewidth, trim={0px, 0px, 100px, 0px}, clip]{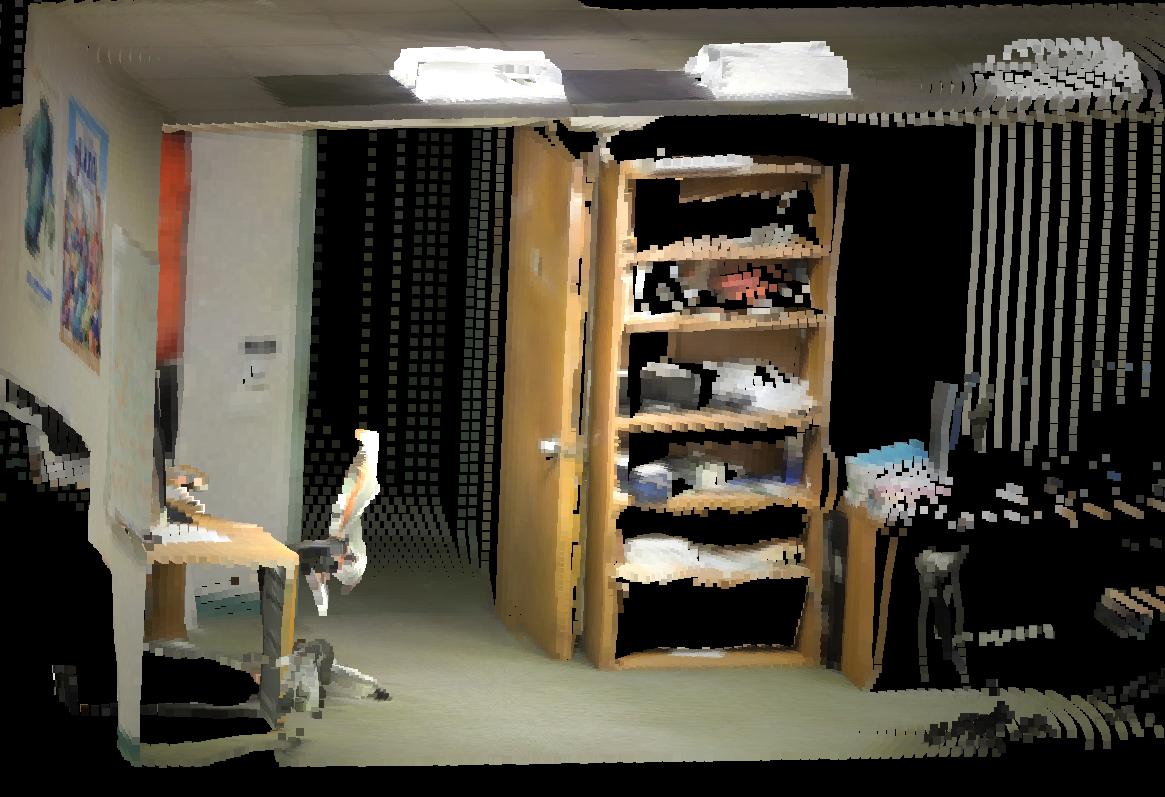}         \\
    \small{BiFuse \cite{9157424}} & \small{Ours} & \small{Ground Truth}\\
    
\end{tabular}
\end{center}
  
  \caption{We propose a global-to-local approach (\emph{i.e.}, GLPanoDepth) to recover a dense depth map from a single panorama. Benefited from the rich global information that have been extracted, the proposed method is able to preserve both global structures and local details as shown in 2D depth maps (top row) and 3D point clouds (bottom row), outperforming previous methods (\emph{e.g.}, BiFuse \cite{9157424}).}
  \label{fig:teaser}
\end{figure}

With the emergence of consumer-level omnidirectional (360$^\circ$) cameras (\emph{e.g.}, Ricoh Theta, Samsung Gear360 and Insta360 ONE), 360$^\circ$ panoramas are currently experiencing a surge in popularity, making depth estimation from these panoramas a hot topic both in academia and in industry \cite{3D60,8885706,8885718,9157424,9157410,10.1007/978-3-030-58517-4_39,Wang_2019_ICRA,9320332,9353978,Sun_2021_CVPR,Pintore_2021_CVPR}. However, directly applying existing deep neural networks trained on perspective images to the omnidirectional domain has been shown to achieve sub-optimal performance due to the space-varying distortion of equirectangular projection \cite{10.1007/978-3-030-01270-0_43,3D60}. Consequently, 
some methods try to reduce the distortion by specially-designed convolution operations \cite{10.1007/978-3-030-01270-0_43} or by combining features from distortion-free cubemaps \cite{9157424,9353978}. Moreover, compared with perspective images, 360$^\circ$ panoramas provide a rich source of information about the entire scene around the viewer. Many neural networks fail to capture the rich features \cite{10.1007/978-3-319-10599-4_43} that are beneficial for high-quality depth estimation. A recent neural network named SliceNet \cite{Pintore_2021_CVPR} tries to exploit the full-geometric context involved in a single panorama. Slicing is performed over multiple levels to preserve global information in the panorama. However, it assumes that the equirectangular image should be aligned to the gravity vector, which does not always hold in practice.

In this paper, we tend to address the above two issues (\emph{i.e.} space-varying distortion and rich global information) simultaneously for panoramic depth estimation, avoiding any strong assumptions on the scenes. Since fully-convolutional networks have a limited receptive field which grows linearly with the depth of the network, they will fail to capture long-range relations and global features that are rich in panoramas \cite{10.1007/978-3-319-10599-4_43}. Considering this, we resort to the vision transformer (ViT) \cite{vit}, a new backbone architecture that is particularly effective at modeling long-range dependencies over all pixels in the input image. Generally, ViT has a global receptive field at every stage. However, directly applying ViT for panoramic depth estimation does not result in higher accuracy than equal-sized convolutional neural networks (CNN) since ViT lacks 2D local structures, \emph{i.e.}, spatial correlations between neighboring pixels. Moreover, ViT also fails to handle distortion in omnidirectional images. Considering these two issues, we design a two-stream network dubbed \emph{GLPanoDepth} which incorporates two branches to extract global and local features, respectively. First, we adopt cubemap projection and construct a new \emph{Cubemap Vision Transformers} (CViT) to model long-range dependencies in spherical signals and extract distortion-free global features. Then, a fully convolutional branch is employed to capture strong 2D local structures that are not well handled by pure transformers. Finally, these two features are fused at multiple scale using a \emph{Gated Fusion Module} to produce high-quality depth maps. 

We have tested our method on multiple datasets. Comparison with the state-of-the-art approaches and extensive ablation studies validate the effectiveness of the proposed method.

\section{Related Work}
The goal in monocular depth estimation is to predict the depth value of each pixel, given only a single RGB image as input. There has been much work which contributes to this field. In this section, we will review previous methods that closely related to our work.

\subsection{Depth Estimation for Perspective Images}
One of the first learning-based methods in monocular depth estimation for perspective images is proposed by Saxena \emph{et al.} \cite{NIPS2005_17d8da81}, using a patch-based model and Markov random fields. After that, many approaches have been presented using hand crafted representations \cite{10518975,7780809,8237759}. With the rapid development of deep learning, the performance of depth estimation from a single image has been significant improvement. Eigen \emph{et al.} \cite{Eigen_2014_NIPS} first applied convolutional neural networks to monocular depth estimation. Following this work, monocular depth estimation based on depth learning has gradually become a research hotspot. Eigen \emph{et al.} \cite{Eigen_2015_ICCV} extended their work \cite{Eigen_2014_NIPS} with a deeper and more discriminative model, using VGG features. Lately, Laina \emph{et al.} \cite{Laina_2016_3DV} proposed FCRN which uses ResNet \cite{He_2016_CVPR} as the encoder and utilizes an up-projection module for upsampling, along with the reverse Huber loss \cite{RePEc:taf:gnstxx:v:28:y:2016:i:3:p:487-514}. Lee \emph{et al.} \cite{Lee_2018_CVPR} also designed a ResNet-based depth estimation network, and resorted to Fourier analysis to well solve the single-image depth estimation problem. As a useful tool, conditional random fields (CRF) have shown great power in optimizing depth estimation results \cite{Liu_2015_CVPR,Cao_2018_TCSVT,Wang_2015_CVPR,Xu_2018_CVPR}. To ease the requirement of ground-truth depth images, unsupervised training methods are developed \cite{Godard_2017_CVPR,Zhan_2018_CVPR,Zhou_2017_CVPR,Yin_2018_CVPR}. However, unsupervised training methods often achieve sup-optimal results compared with supervised training methods. Similar to ours, Ranftl \emph{et al.} \cite{Ranftl2021} also tried to recover dense depth values from a single image by vision transformers. However, they only focused on distortion-free perspective images.
For a comprehensive review on depth estimation for perspective images, please refer to some recent surveys \cite{DBLP:journals/corr/abs-1906-06113,s20082272,MING202114}.

\subsection{Depth Estimation for Omnidirectional Images}
Perspective images contain limited geometric contexts because of the restricted field-of-view \cite{10.1007/978-3-319-10599-4_43}. With the popularity of omnidirectional cameras, researches on depth estimation for omnidirectional images have emerged \cite{Wang_2019_ICRA,Zou_2018_CVPR,10.1007/978-3-319-10599-4_43}. It has been shown that directly applying existing deep learning methods designed for perspective images to the omnidirectional domain achieves sub-optimal results.
To deal with the distortion of equirectangular projection in omnidirectional images, some methods utilize cubemaps as an additional cue. Cheng \emph{et al.} \cite{Cheng_2018_CVPR} first suggested to convert an equirectangular image into a cubemap to address the issue of space-varying distortion, while keeping the connection between each adjacent face of the cubemap. Wang \emph{et al.} \cite{Wang_2018_ACCV} also adopted the cubemap representation in a self-supervised learning manner. To properly combine features from different representations of spherical signals, specially designed fusion strategies are proposed \cite{9157424,9353978}. Another way to handle distortion is by specially-designed spherical convolutions \cite{NIPS2017_0c74b7f7,Su_2019_CVPR,s.2018spherical,10.1007/978-3-030-01240-3_32}. Several methods use them to make the networks explicitly aware of the distortion. By incorporating spherical convolutions \cite{NIPS2017_0c74b7f7}, Zioulis \emph{et al.} \cite{3D60} utilized omnidirectional content in the form of equirectangular images to perform depth estimation. Most recently, Pintore \emph{et al.} \cite{Pintore_2021_CVPR} achieved remarkable results using a slice-based representation of omnidirectional images. They introduced a LSTM multi-layer module \cite{10.5555/2969239.2969329} to effectively recover long and short term spatial relationships between slices.

\section{Method}
\begin{figure*}[t]
  \centering
  
   \includegraphics[width=1.0\linewidth]{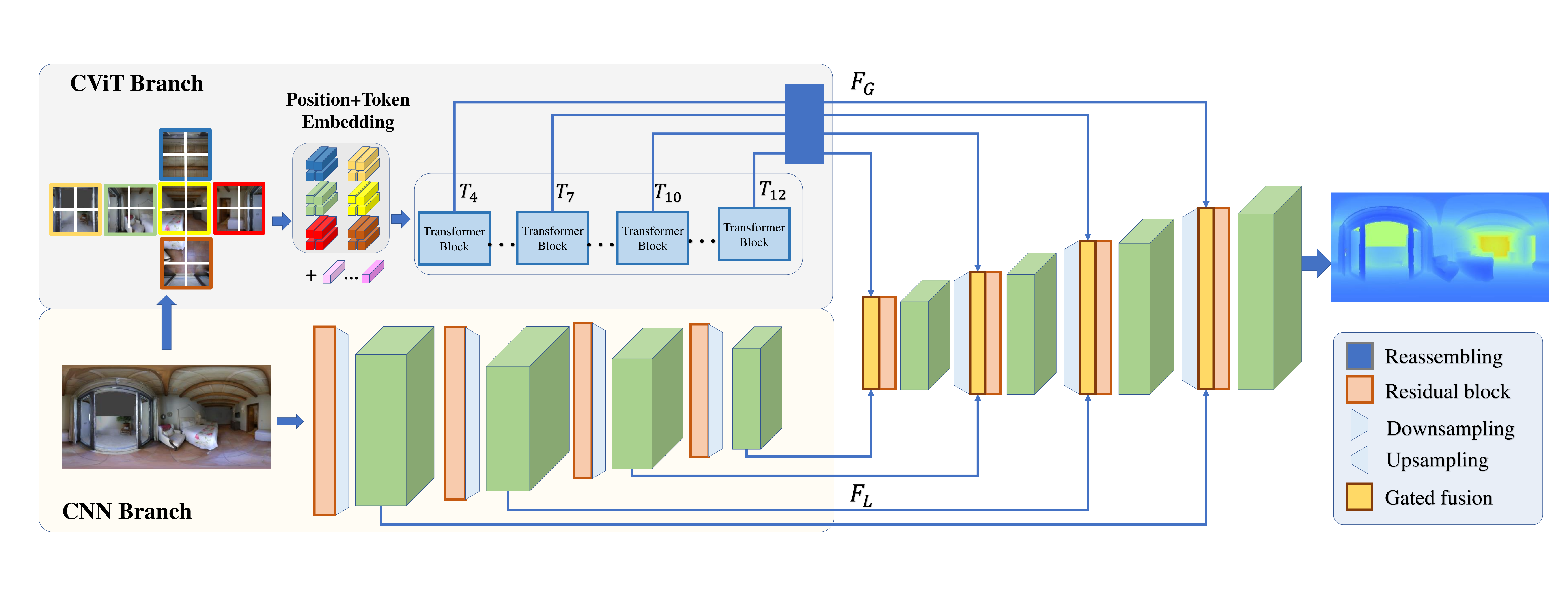}

   \caption{The overall architecture of GLPanoDepth. In the CViT branch, a cubemap is transformed into tokens by extracting non-overlapping patches on each face, followed by a linear projection of their flattened representation. This image embedding is augmented with a positional embedding and is passed to multiple transformer blocks. After reassembling tokens from different blocks, we obtain global features $\mathcal{F}_{G}$ at multiple scales. In the CNN branch, residual blocks are applied to extract hierarchical local features $\mathcal{F}_{L}$. A gated fusion module is designed to fuse features from two branches, leading to the final prediction.}
   \label{fig:network}
\end{figure*}

\subsection{Motivation and Overview}
Our goal is to predict a dense depth map from an omnidirectional image. With the advent of practical deep learning, many methods apply convolutional neural networks to extract and decode the features from the input image.
Unfortunately, these convolutional neural networks are not well suited for omnidirectional or spherical images. 
Convolutions are linear operations that have a limited receptive field. They can hardly acquire sufficiently broad context as well as significant high-level representations from omnidirectional images. The limited receptive field and the limited expressiveness of an individual convolution operation necessitate sequential stacking many convolutional layers into very deep architectures. 
In contrast, vision transformers \cite{vit} outperform CNNs in long-range feature learning. However, they have difficulty in learning local structures especially for panoramas with heavy space-varying distortion.

In view of these challenges, we design a two-stream network, named GLPanoDepth, for panoramic depth estimation. An overview of the complete architecture is shown in Fig.~\ref{fig:network}. The key idea of our method is to extract global and local features with two separate branches, which leads to fine-grained and globally coherent prediction. To extract global features, we employ cubemap vision transformers (CViT), a new transformer-based architecture which learns long-range contextual relationships and distortion-free global features through multi-head self-attention (MHSA) \cite{NIPS2017_3f5ee243} and multi-layer perceptrons (MLP). As transformers are set-to-set functions, they do not perform well in retaining the information of the spatial positions and local features. To address this issue, we leverage a convolution-based subnetwork to learn highly correlated local structures from panoramas. Finally, a gated fusion module is designed to progressively combine features from two branches into the final dense prediction. 

\subsection{Network Architecture}
As shown in Fig.~\ref{fig:network}, our network can be logically separated to an encoder part and a decoder part. The encoder consists of a CViT branch and a CNN branch, both of which provide features at various resolutions. Given an input panorama $\mathcal{P}$ with a $H\times W$ resolution, the CViT branch is designed to capture global features $\mathcal{F}_{G}$ while the CNN branch is expected to learn local features $\mathcal{F}_{L}$. The decoder then progressively combine these features into a full resolution prediction using the gated fusion module. 

\textbf{CViT Branch }
We leverage vision transformers \cite{vit} as the backbone of CViT branch. Vision transformers are believed to capture global representations among the compressed patch embeddings with dynamic attention, global context fusion and better generation. Due to the distortion in omnidirectional images, our CViT branch takes the cubemap projection as the input. Specifically, we first reproject the input panorama $\mathcal{P}$ into a corresponding cubemap $\mathcal{C}\in \mathbb{R}^{6\times \frac{H}{2}\times \frac{H}{2}\times 3}$, where 6 means six faces corresponding to the planes on the back, down, front, left, right and up. Note that the image of each face is distortion-free while all the faces form a complete 360$^\circ$ field-of-view.
Then, we split the images from the six faces into a sequence of flattened patches of size $p\times p$ pixels. All these non-overlapping square patches are flattened into vectors and individually embedded using a linear projection, resulting in $N_p=(6\times \frac{H}{2}\times  \frac{H}{2})/p^2$ tokens in total.

Since we have lost the information of the spatial positions of individual tokens during flatting, we add a learnable position embedding to the image embeddings.
After that, input tokens are transformed into multi-scale representations ${T_k}$ using $K$ transformer blocks. ${T_k}$ is the output of $k$-th transformer block. 
Fundamentally, each transformer block contains two basic operations. First, an attention operation is adopted to model inter-element relations. Second, an MLP is used to model relations within each element. These two operations are often intertwined with normalization and residual connections. Since each transformer block is performed on tokens from all six faces, it naturally addresses the discontinuity issue of cubemap projection.

To reassemble tokens from different transform blocks, we map the set of tokens into image-like feature representations. Formally, we apply a linear projection and spatial reshape operation to adjust $N_p$ $C$-channel tokens $T_k \in \mathbb{R}^{N_p\times C}$ into a feature map $\mathcal{F}_{T_k}\in \mathbb{R}^ {\frac{H}{p}\times \frac{W}{p} \times C}$. Since the spatial reshape operation has limited information for scales, we add a recover layer to scale the representation. This layer is implemented using $1\times 1$ convolutions and pixelshuffle~\cite{pixelshuffle}. Through this layer, feature maps from earlier blocks have higher resolutions and have a global receptive field. Currently, we use 12 transformer blocks, \emph{i.e.}, $K=12$, in the CViT branch. We only extract global feature maps from the first, fourth, seventh and twelfth transformer blocks, \emph{i.e.},
\begin{equation}
    \mathcal{F}_{G} = \{\mathcal{F}_{T_{4}},\mathcal{F}_{T_{7}}, \mathcal{F}_{T_{10}}, \mathcal{F}_{T_{12}}\}.
\end{equation}
These feature maps will be fed into the decoder.

\textbf{CNN Branch }
The CNN branch is devoted to capture local relationships using local receptive fields and shared weights in CNN. It takes the panorama $\mathcal{P}$ with equirectangular projection as the input and outputs local feature maps $\mathcal{F}_{L}$ with the same size as those from the CViT branch. Currently, we do not choose any spherical variant of convolution to handle distortion in the panorama because our ultimate goal is to produce an equirectangular depth map. The CViT branch is responsible for correcting distorted regions.

In the CNN branch, we have designed four residual blocks\cite{He_2016_CVPR} and downsampling layers to extract local features from detail-rich shallow layers and context-critical deep layers. In the residual block, two $3\times 3$ convolutional layers are used to address limited-receptive-field learning for specific local details, each followed by a rectified linear unit (ReLU). We apply a $3\times 3$ convolutional layer with stride 2 for downsampling.

Our CNN branch extracts a hierarchical ensemble of local features with multiple limited reception fields from different layers. These features, denoted as $\mathcal{F}_{L}$, have the same resolution as $\mathcal{F}_{G}$ from the CViT branch.

\begin{figure}[t]
  \centering
   \includegraphics[width=1.0\linewidth]{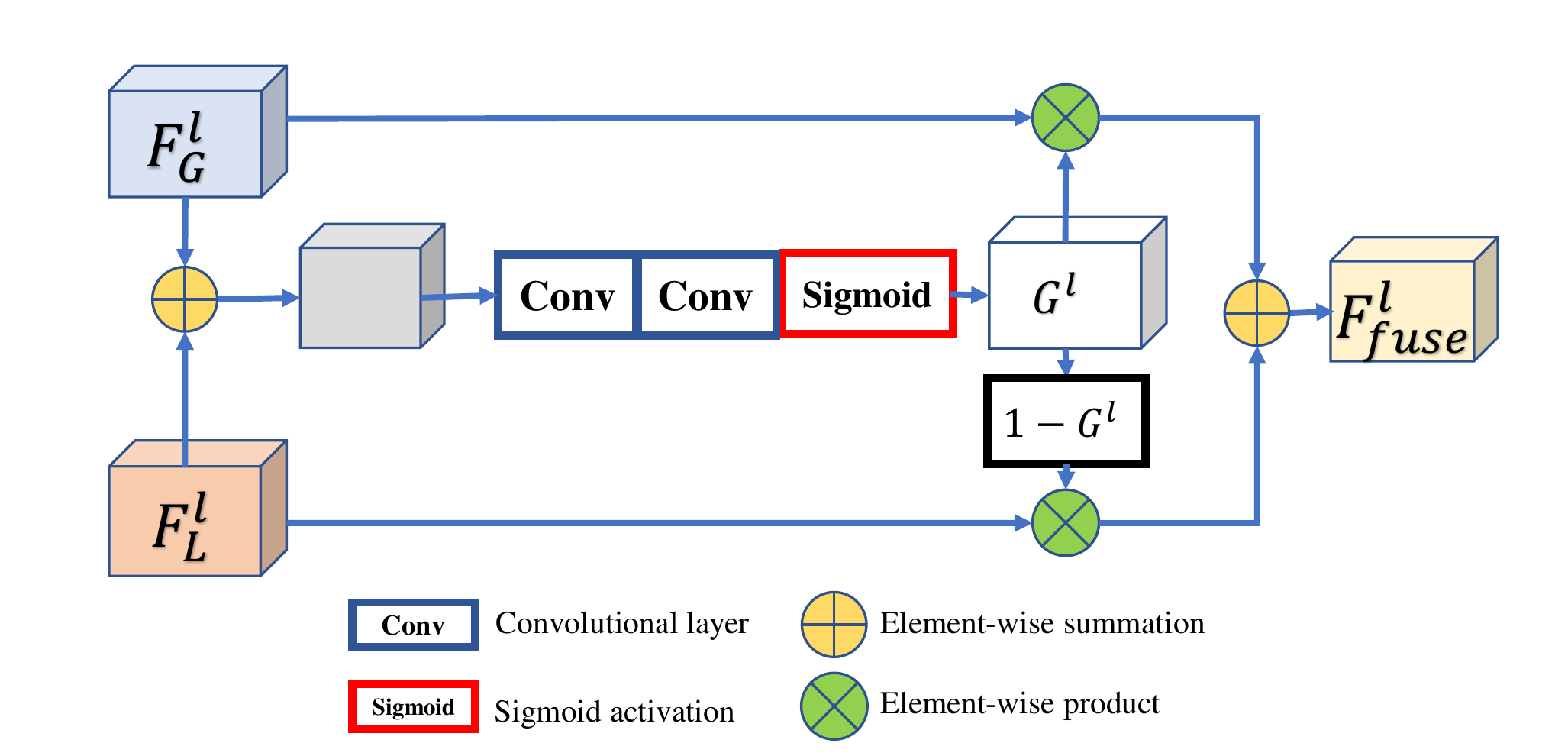}
   \caption{Illustration of the proposed gated fusion module. 
   }
   \label{fig:fuse_module}
\end{figure}

\textbf{Gated Fusion Module }
To combine information from two branches, we propose a gated fusion module to adaptively adjust the weights assigned to global and local features. As illustrated in Fig.~\ref{fig:fuse_module}, we first fuse feature maps ($\mathcal{F}^l_{G}$ and  $\mathcal{F}^l_{L}$) from $l$-th layer of the two branches via an element-wise summation, and then pass the result to two convolutional layers and a sigmoid function to generate a gate map $G^l$:
\begin{equation}
G^l = \text{Sigmoid}(\text{Conv}(\text{Conv}(\mathcal{F}^l_{G} +  \mathcal{F}^l_{L}))).
\end{equation}
$G^l$ ranges from 0 to 1, serving as a gate that select different features from the two branches. We use $G^l$ and $1-G^l$ as soft attention matrices for $\mathcal{F}_{G}^l$ and $ \mathcal{F}_{L}^l$, respectively. Then, we obtain a fused feature map $\mathcal{F}_{fuse}^l$ as
\begin{equation}
\mathcal{F}^l_{fuse} = \mathcal{F}^l_{G}\otimes G^l + \mathcal{F}^l_{L} \otimes(1-G^l)
\end{equation}
in which $\otimes$ denotes element-wise product.

Our gated fusion module has two fundamental differences against bi-projection fusion module proposed by Wang \emph{et al.} \cite{9157424}. First, we fuse different feature maps into a single decoder---instead of two---to produce an omnidirectional image directly. Second, the feature maps generated by the CViT branch do not require a cubemap to equirectangular transformation and suffer less from inconsistency in cubemap boundaries thanks to the transformer blocks.

\begin{table*}[tp]
  \centering
  \begin{threeparttable}
    \begin{tabular}{clcccccc}
    \toprule
    \bf Dataset & \bf Method & \bf MAE $\downarrow$ &  \bf RMSE $\downarrow$ & \bf RMSElog $\downarrow$ & $\delta\leq 1.25$ $\uparrow$ & $\delta\leq 1.25^{2}$ $\uparrow$ & $\delta\leq 1.25^{3}$ $\uparrow$\cr
    \midrule
        \multirow{8}{*}{\bf 360D}& FCRN~\cite{Laina_2016_3DV}                              &0.1381&0.2833&0.0473&0.9532&0.9905&0.9966\cr
    &OmniDepth~\cite{3D60}&0.1706 &0.3171&0.0725&0.9092&0.9702&0.9851\cr
     &BiFuse~\cite{9157424}&0.1143&0.2440&0.0428&0.9699&0.9927&0.9969\cr
     & SliceNet~\cite{Pintore_2021_CVPR} &0.1134   &\bf{0.1323}&\bf{0.0212}&\underline{0.9788}&0.9952&0.9969\cr
     &Ours(CViT)\tnote{1}      &0.1791  &0.3555   &0.0604   &0.9219   &0.9851   &0.9951\cr
    &Ours(ViT+CNN)                                    &0.1279&0.2316&0.0392&0.9681&0.9946&\underline{0.9983}\cr
    &Ours(Concat)   &\underline{0.1104}   &0.2011   &0.0344 &0.9786  &\underline{0.9961} &\bf{0.9987}\cr
     &Ours(Complete) &{\bf 0.0978}&\underline{0.1844}&\underline{0.0313}&{\bf 0.9834}&{\bf 0.9965} & {\bf 0.9987}\cr
     \midrule
    \multirow{7}{*}{\bf Matterport3D}& FCRN~\cite{Laina_2016_3DV}                              &0.4008&0.6704&0.1244&0.7703&0.9174&0.9617\cr
    &OmniDepth~\cite{3D60}&0.4838   &0.7643&0.1450&0.6830&0.8794&0.9429\cr
     & BiFuse~\cite{9157424}             &0.3470&0.6259&0.1134&0.8452&0.9319&0.9632\cr
     & SliceNet~\cite{Pintore_2021_CVPR} &0.3296&0.6133&0.1045&\bf0.8716&0.9483&0.9716\cr
     &Ours(ViT+CNN)   &\underline{0.3232}    &\underline{0.5466}    &\underline{0.0827}    &0.8476   &\underline{0.9512}   &\underline{0.9793}\cr
    &Ours(Concat)     &0.3407    &0.5747   &0.0862    &0.8288   &0.9466   &0.9779\cr
     &Ours(Complete) &\bf0.2998&\bf 0.5223&\bf 0.0786& \underline{0.8641}&\bf 0.9561&\bf 0.9808\cr
     
    \midrule
    \multirow{7}{*}{\bf Stanford2D3D}& FCRN~\cite{Laina_2016_3DV}                              &0.3428&0.4142&0.1100&0.7230&0.9207&0.9731\cr
    &OmniDepth~\cite{3D60}&0.3743 &0.6152&0.1212&0.6877&0.8891&0.9578\cr
    &BiFuse~\cite{9157424} &0.2343&0.4142&0.0787&0.8660&0.9580&\underline{0.9860}\cr
    & SliceNet\tnote{2}~\cite{Pintore_2021_CVPR}&\bf 0.1715&\bf 0.3383& \underline{0.0752}&\bf 0.9059&\underline{0.9635}&0.9848\cr
    &Ours(ViT+CNN)  &0.2739    &0.4810    &0.0901    &0.8172   &0.9386   &0.9773\cr
    &Ours(Concat)   &0.2734    &0.4829    &0.0906    &0.8211   &0.9416   &0.9762\cr
     &Ours(Complete) &\underline{0.1932}&\underline{ 0.3493}&\bf{0.06801}&\underline{ 0.9015}&\bf{ 0.9739}&\bf{0.9901}\cr
     
    \bottomrule
    \end{tabular}
    \begin{tablenotes}
        \footnotesize
        \item[1] Since both Matterport3D and Stanford2D3D contain large areas of missing depth value, using a single CViT branch fail to converge on these two datasets. Therefore, we only provide numerical metrics of 360D for Ours(CViT).
        \item[2] These numerical metrics are gained via re-evaluating the pre-trained model released on the project's website (https://github.com/crs4/SliceNet).
      \end{tablenotes}
    \end{threeparttable}
      \caption{Quantitative performance on three benchmark datasets. The best results are highlighted in \textbf{bold}, and the second are \underline{underlined}.}
  \label{tab:performance_comparison}
\end{table*}

\subsection{Loss Function and Training Details}
Following most recent works on panoramic depth estimation, we adopt the BerHu loss~\cite{Laina_2016_3DV} as the objective function in training:
\begin{equation}
L(y,\hat{y})=\left\{
\begin{aligned}
 &|y-\hat{y}|, & |y-\hat{y}| \leq T \\
&\frac{(y-\hat{y})^2+T^2}{2T},  & |y-\hat{y}| \ge T
\end{aligned}
\right.
\label{eq6}
\end{equation}
where $y$ is the ground truth depth value and $\hat{y}$ is the prediction. The threshold $T$ is set to 0.2 in our experiments.

We implement our GLPanoDepth using the PyTorch framework~\cite{pytorch}. Adam~\cite{kingma2017adam} optimizer is used with the default parameters $\beta_1=0.9$ and $\beta_2 =  0.999$ and an initial learning rate of 0.0001.
Our model is trained on two NVIDIA RTX 3090 GPUs with a batch size of 8. We jointly train two branches for 80 epochs. We'll release the code and trained models upon acceptance.

\section{Experiments}
In this section, we present extensive experimental results to validate the proposed method. We report both quantitative and qualitative results of three large-scale datasets and compare our method with state of the arts. Ablation studies are carried out to validate our design choices. More results are provided in the supplemental material.

\begin{figure*}[tbp]
  \begin{center}
  \renewcommand\tabcolsep{1.0pt}
  \begin{tabular}{ccccc}
    \includegraphics[width=0.2\linewidth, trim={0px, 0px, 0px, 0px}, clip]{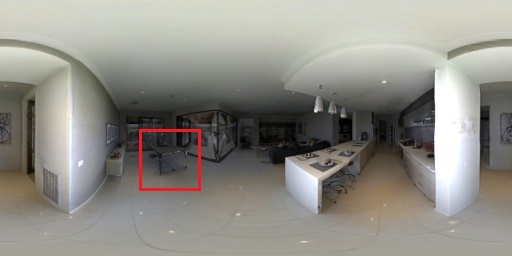}           &
    \includegraphics[width=0.2\linewidth, trim={0px, 0px, 0px, 0px}, clip]{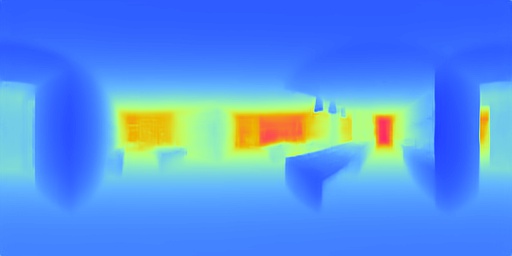}      &
    \includegraphics[width=0.2\linewidth, trim={0px, 0px, 0px, 0px}, clip]{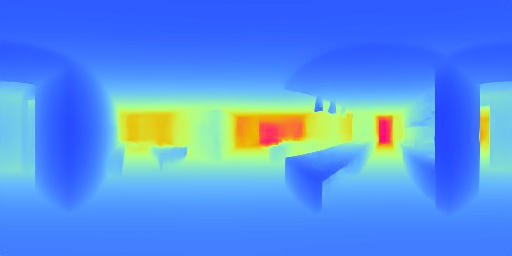}    &
    \includegraphics[width=0.2\linewidth, trim={0px, 0px, 0px, 0px}, clip]{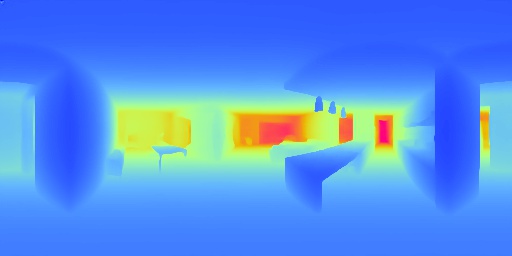}        &
    \includegraphics[width=0.2\linewidth, trim={0px, 0px, 0px, 0px}, clip]{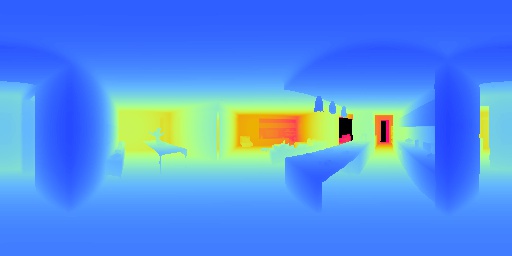}         \\
    
    \includegraphics[width=0.2\linewidth, trim={0px, 0px, 0px, 0px}, clip]{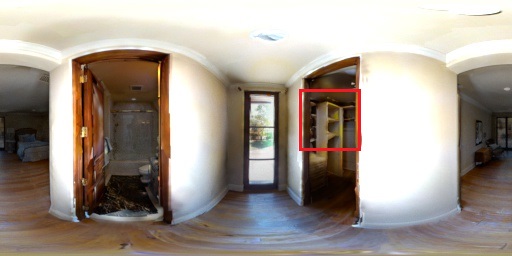}           &
    \includegraphics[width=0.2\linewidth, trim={0px, 0px, 0px, 0px}, clip]{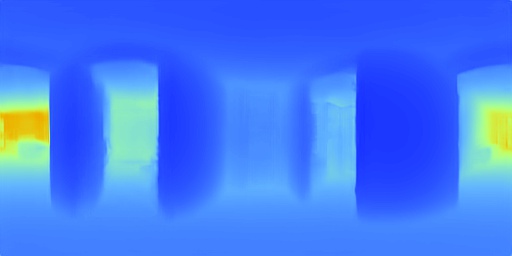}      &
    \includegraphics[width=0.2\linewidth, trim={0px, 0px, 0px, 0px}, clip]{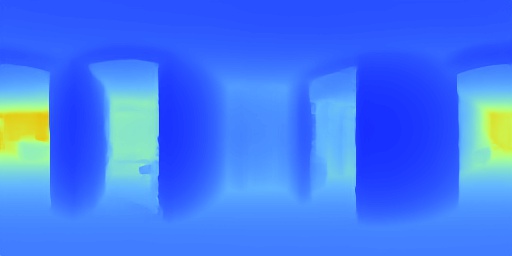}    &
    \includegraphics[width=0.2\linewidth, trim={0px, 0px, 0px, 0px}, clip]{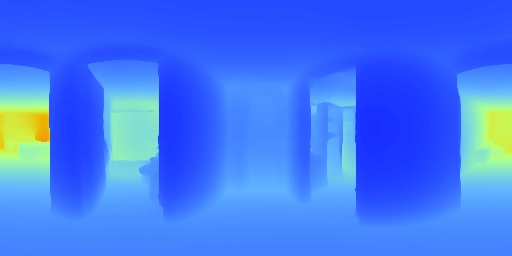}        &
    \includegraphics[width=0.2\linewidth, trim={0px, 0px, 0px, 0px}, clip]{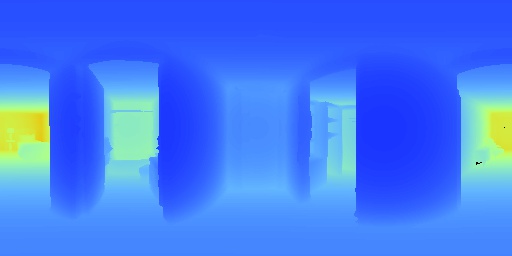}         \\

    \includegraphics[width=0.2\linewidth, trim={0px, 0px, 0px, 0px}, clip]{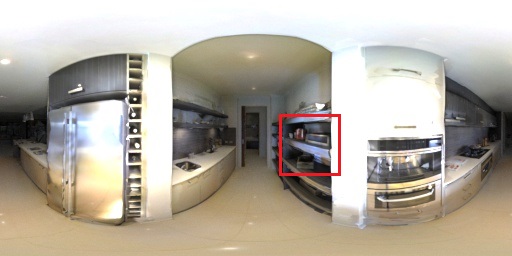}           &
    \includegraphics[width=0.2\linewidth, trim={0px, 0px, 0px, 0px}, clip]{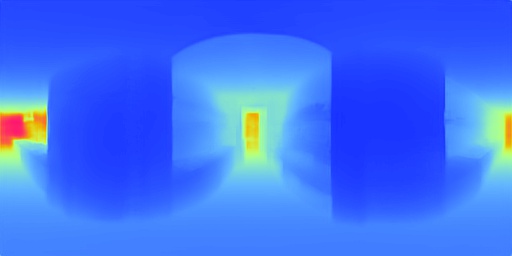}      &
    \includegraphics[width=0.2\linewidth, trim={0px, 0px, 0px, 0px}, clip]{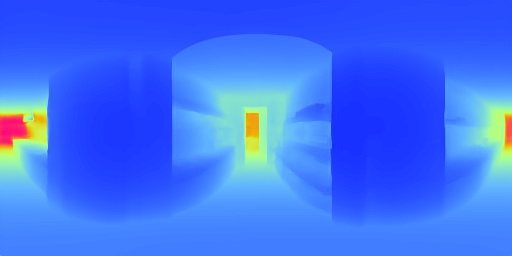}    &
    \includegraphics[width=0.2\linewidth, trim={0px, 0px, 0px, 0px}, clip]{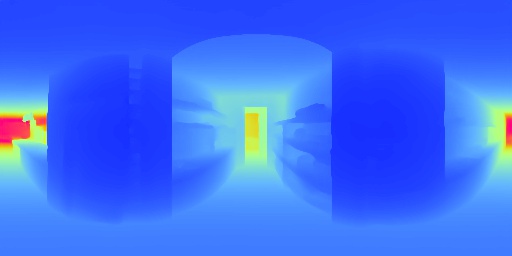}        &
    \includegraphics[width=0.2\linewidth, trim={0px, 0px, 0px, 0px}, clip]{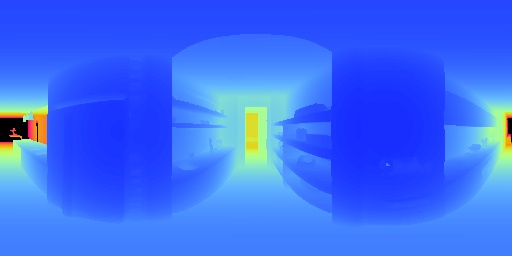}         \\

    \includegraphics[width=0.2\linewidth, trim={0px, 0px, 0px, 0px}, clip]{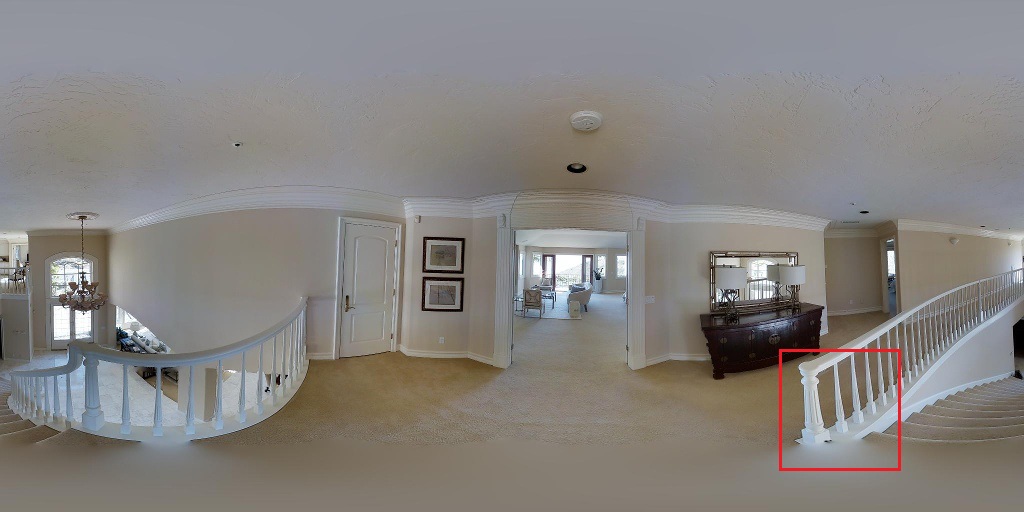}             &
    \includegraphics[width=0.2\linewidth, trim={0px, 0px, 0px, 0px}, clip]{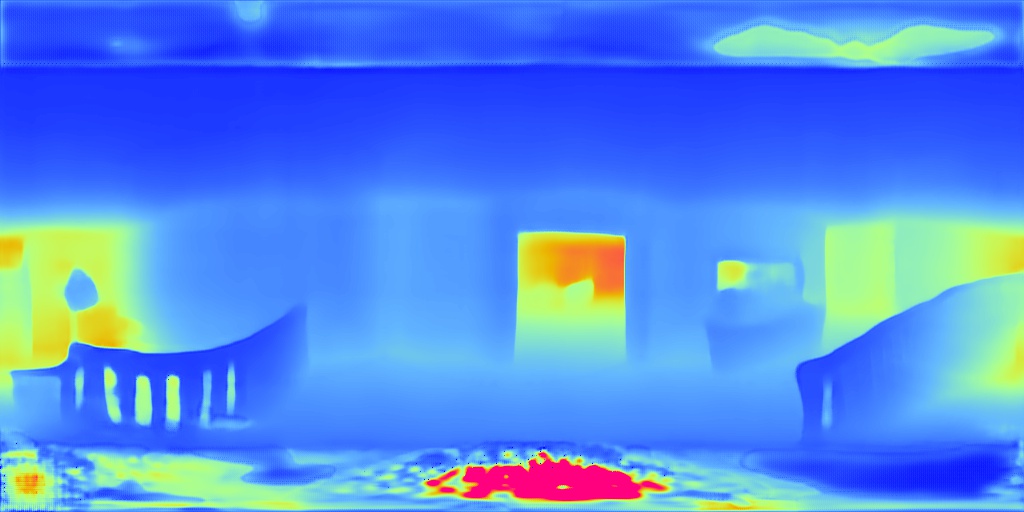}  &
    \includegraphics[width=0.2\linewidth, trim={0px, 0px, 0px, 0px}, clip]{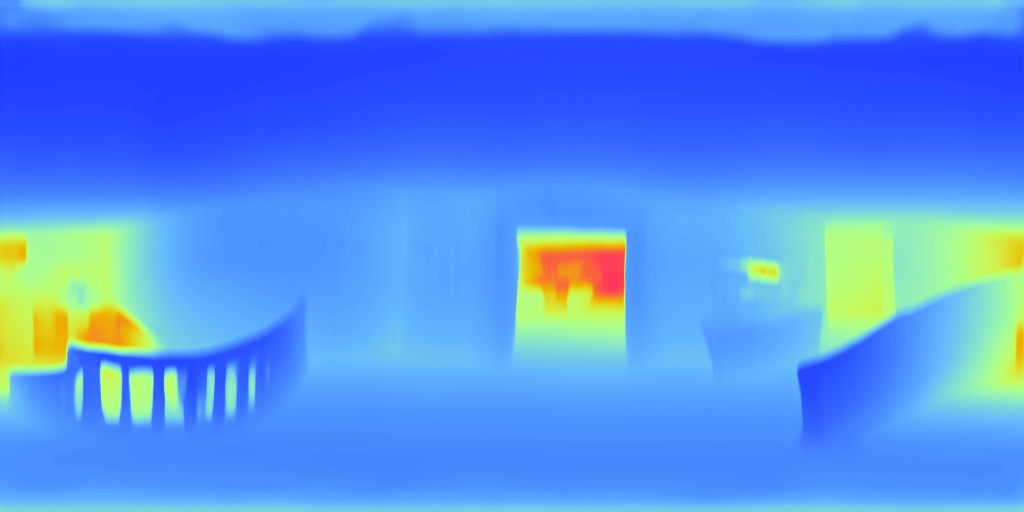} &
    \includegraphics[width=0.2\linewidth, trim={0px, 0px, 0px, 0px}, clip]{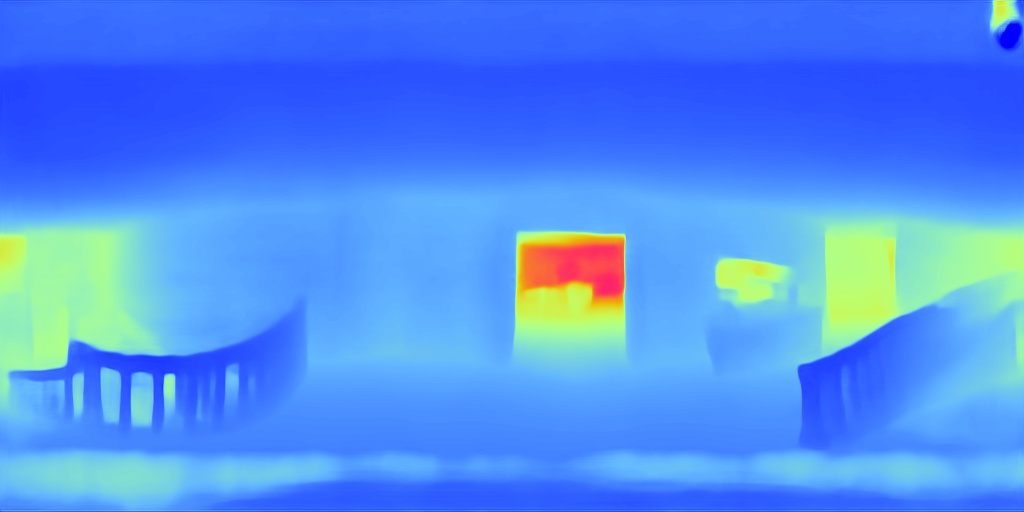} &
    \includegraphics[width=0.2\linewidth, trim={0px, 0px, 0px, 0px}, clip]{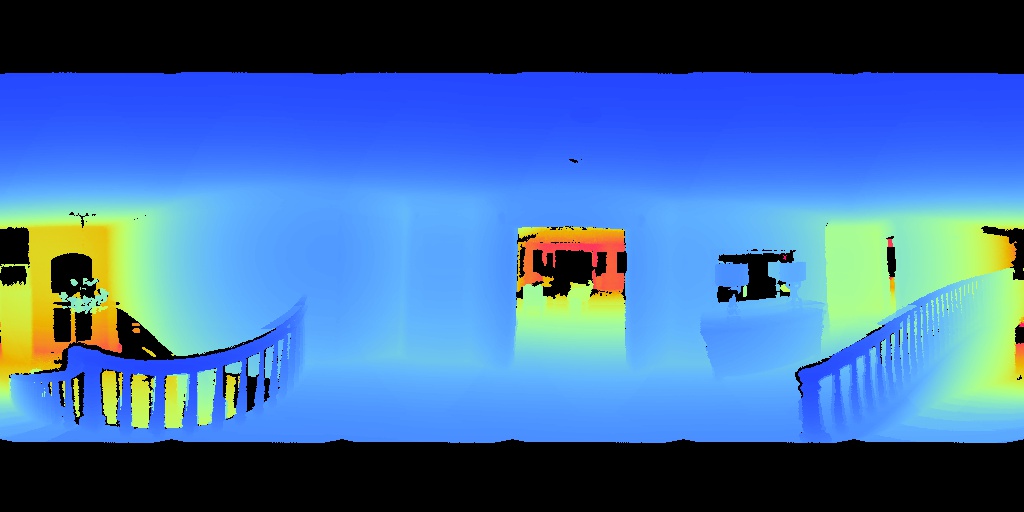}\\
    
    \includegraphics[width=0.2\linewidth, trim={0px, 0px, 0px, 0px}, clip]{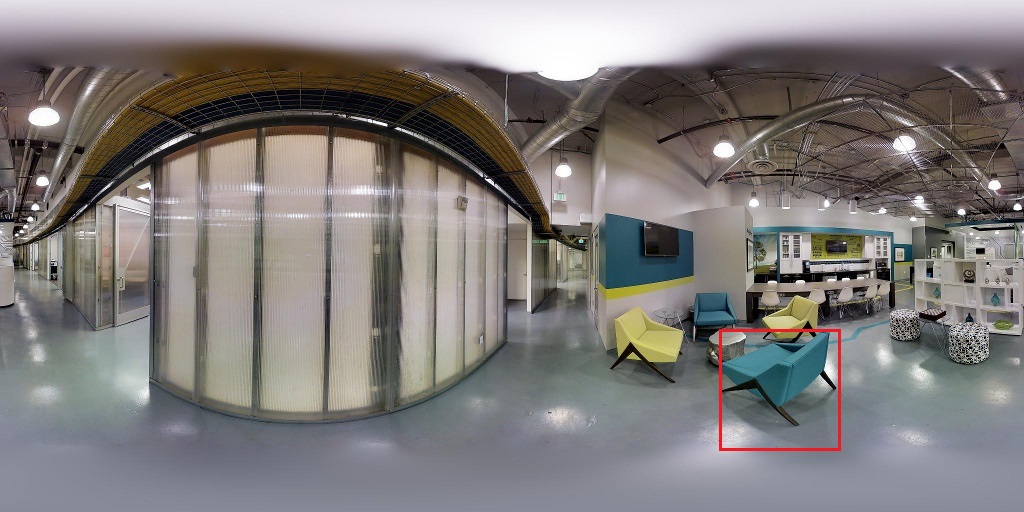}             &
    \includegraphics[width=0.2\linewidth, trim={0px, 0px, 0px, 0px}, clip]{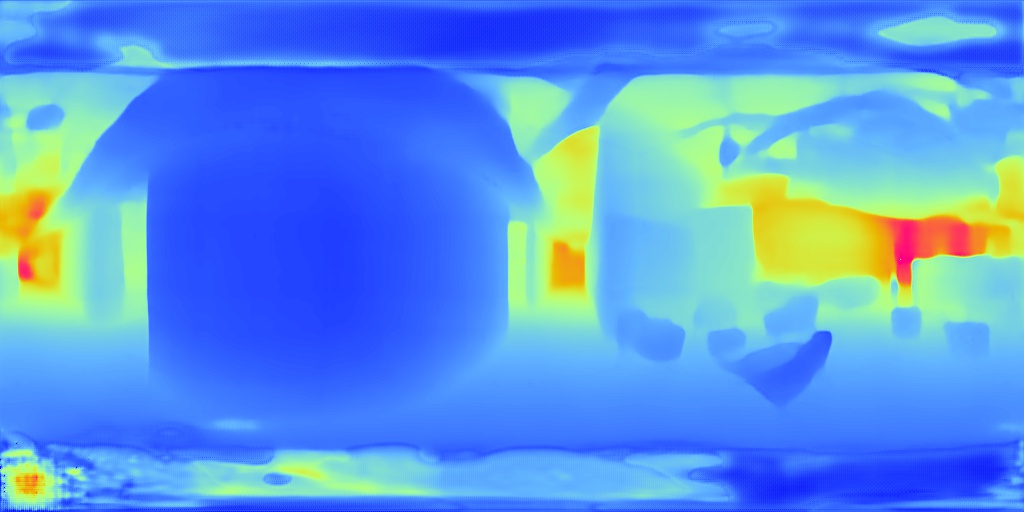}  &
    \includegraphics[width=0.2\linewidth, trim={0px, 0px, 0px, 0px}, clip]{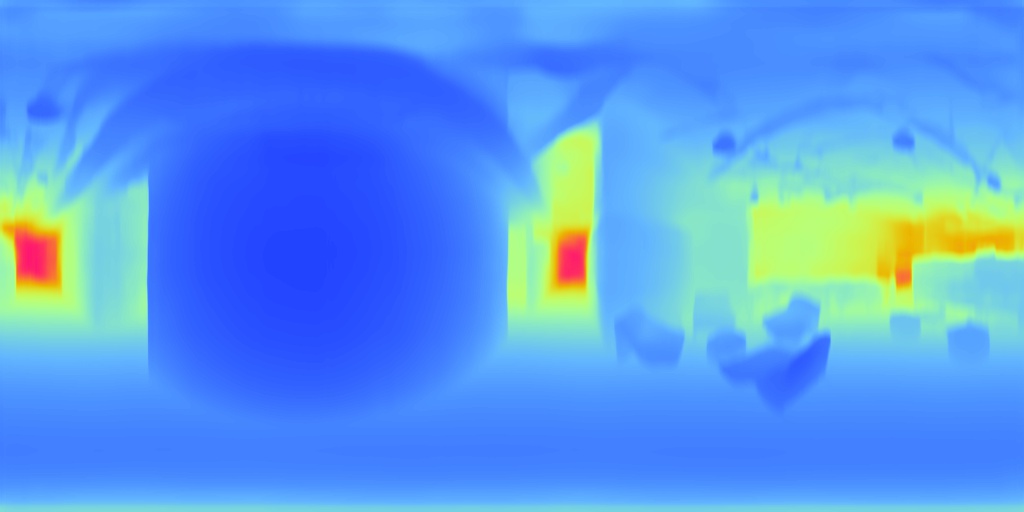} &
    \includegraphics[width=0.2\linewidth, trim={0px, 0px, 0px, 0px}, clip]{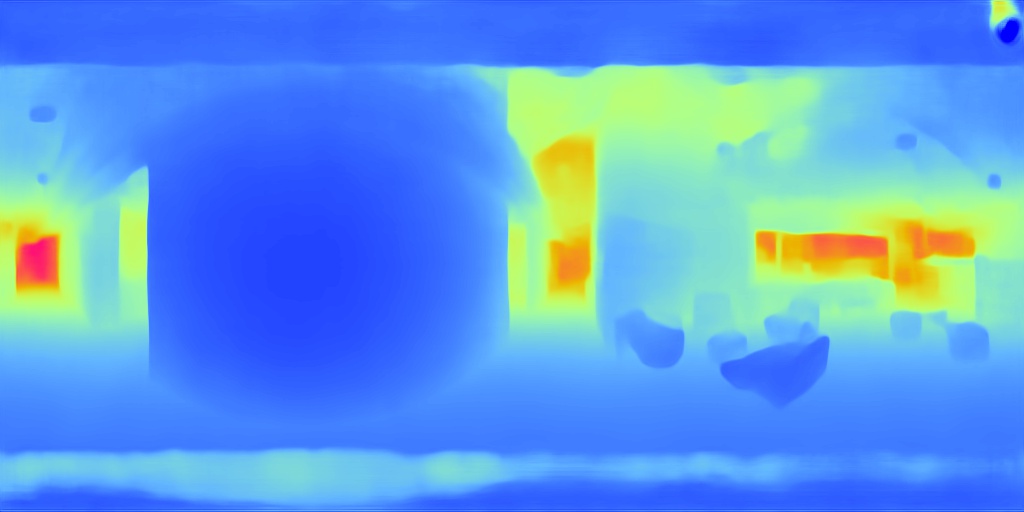} &
    \includegraphics[width=0.2\linewidth, trim={0px, 0px, 0px, 0px}, clip]{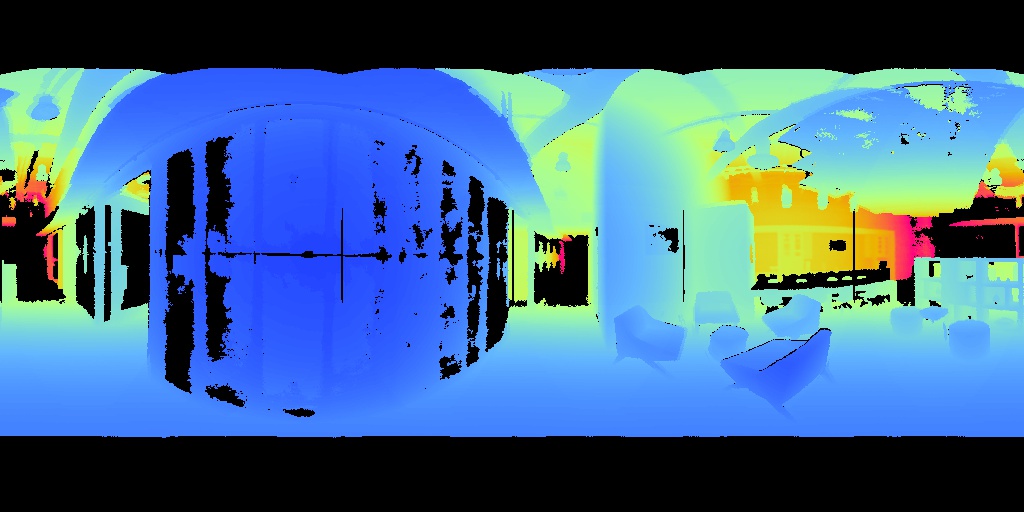}\\

  
    \includegraphics[width=0.2\linewidth, trim={0px, 0px, 0px, 0px}, clip]{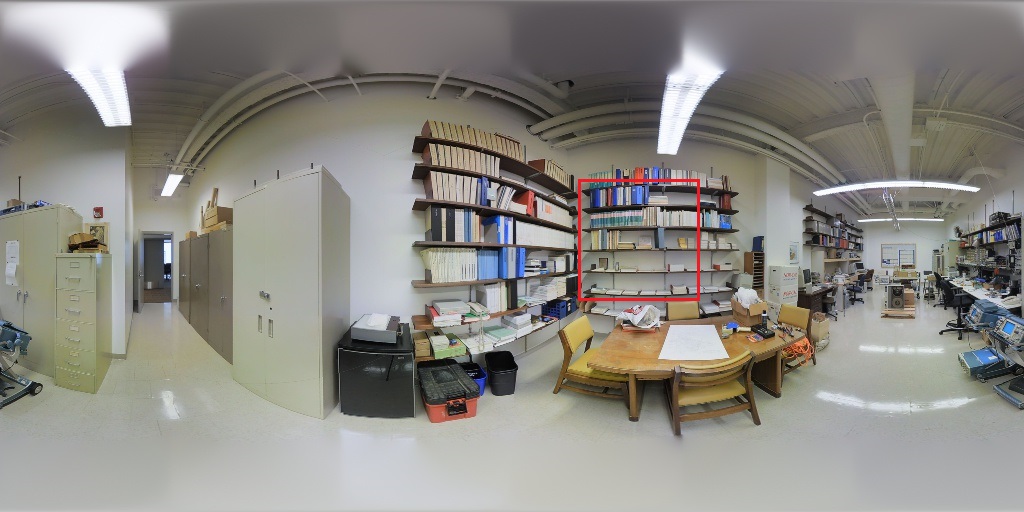}  &
    \includegraphics[width=0.2\linewidth, trim={0px, 0px, 0px, 0px}, clip]{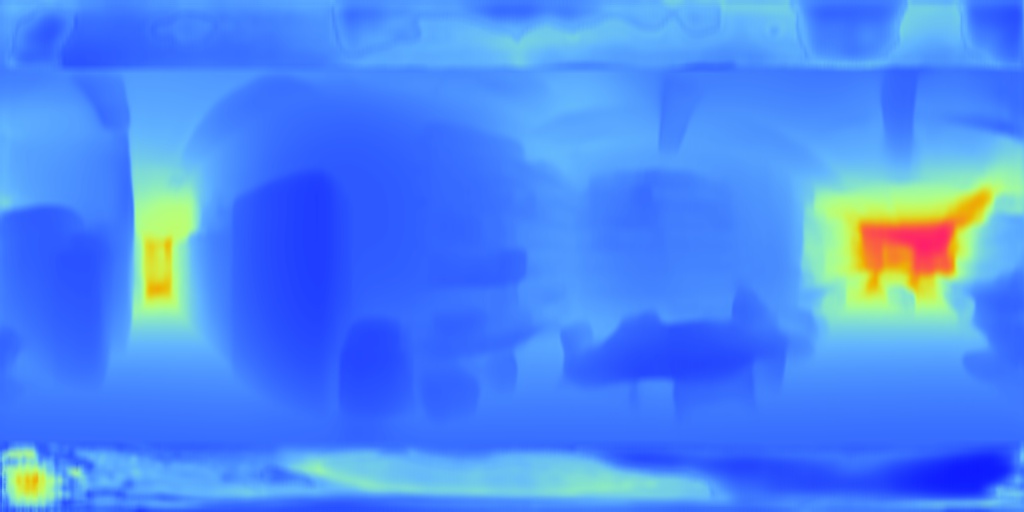}  &
    \includegraphics[width=0.2\linewidth, trim={0px, 0px, 0px, 0px}, clip]{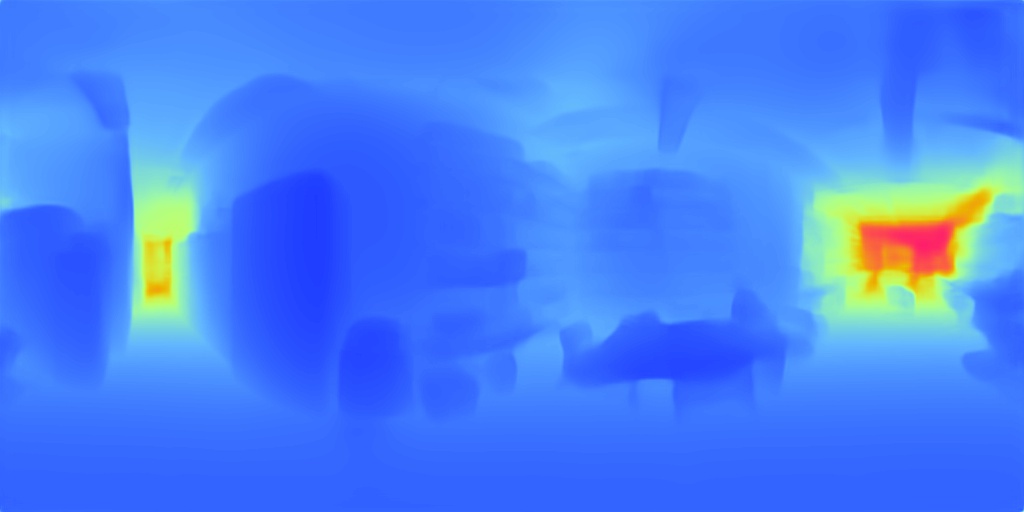} &
    \includegraphics[width=0.2\linewidth, trim={0px, 0px, 0px, 0px}, clip]{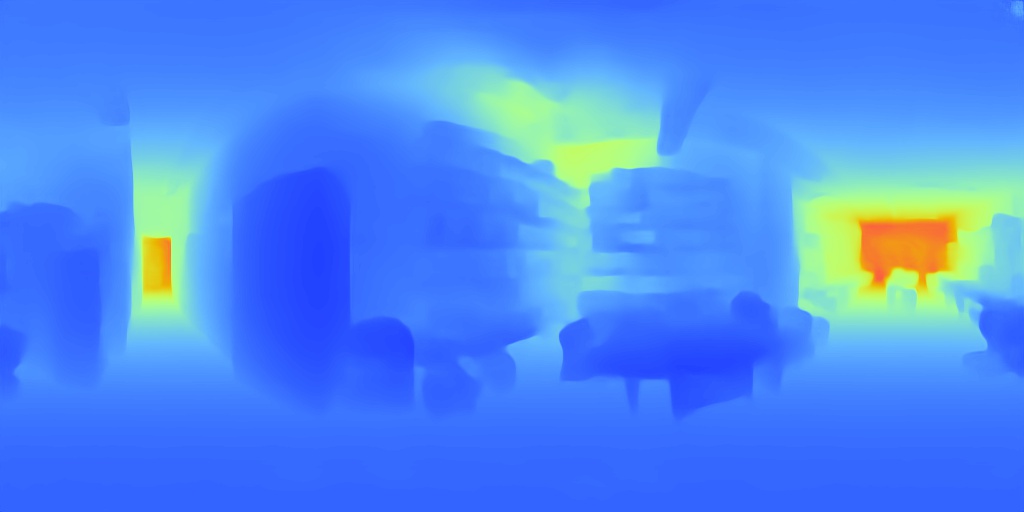} &
    \includegraphics[width=0.2\linewidth, trim={0px, 0px, 0px, 0px}, clip]{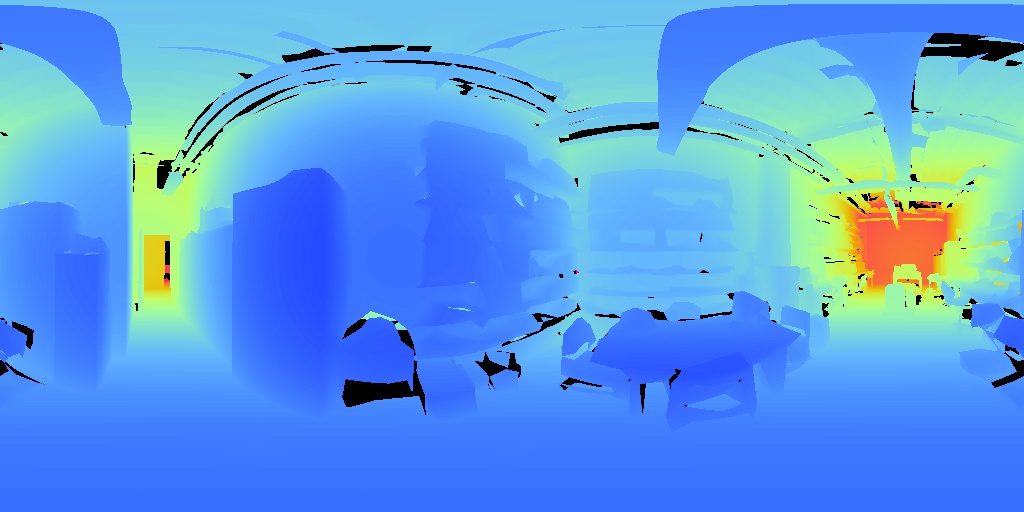}\\
    
    \includegraphics[width=0.2\linewidth, trim={0px, 0px, 0px, 0px}, clip]{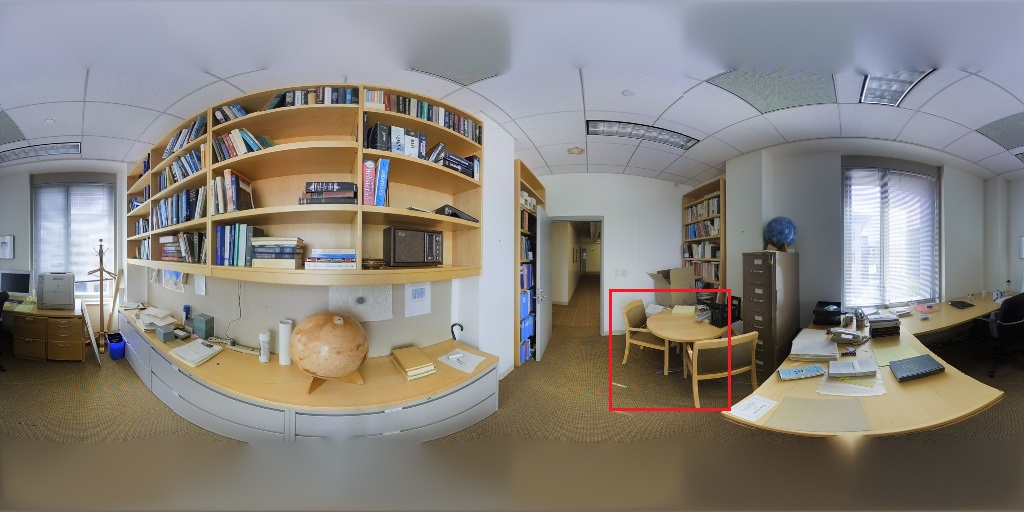}  &
    \includegraphics[width=0.2\linewidth, trim={0px, 0px, 0px, 0px}, clip]{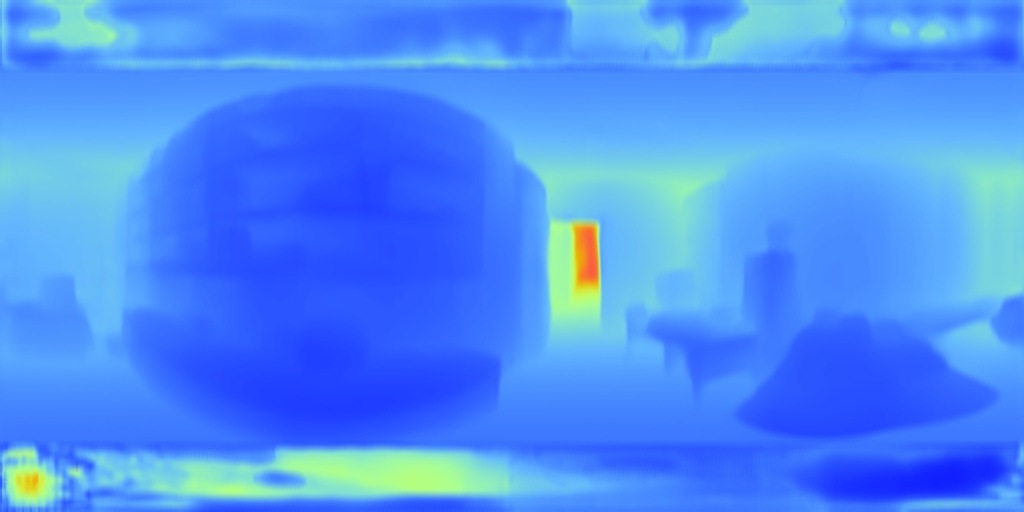}  &
    \includegraphics[width=0.2\linewidth, trim={0px, 0px, 0px, 0px}, clip]{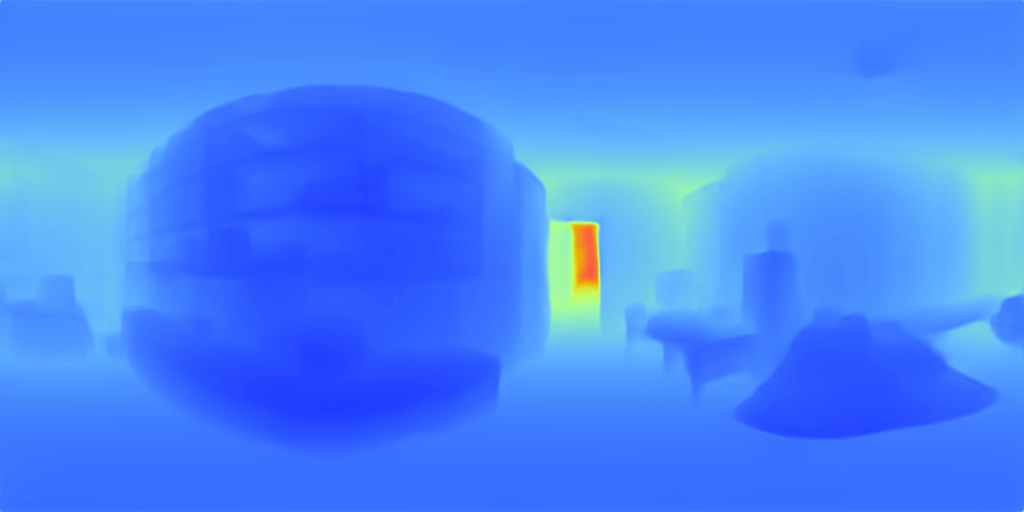} &
    \includegraphics[width=0.2\linewidth, trim={0px, 0px, 0px, 0px}, clip]{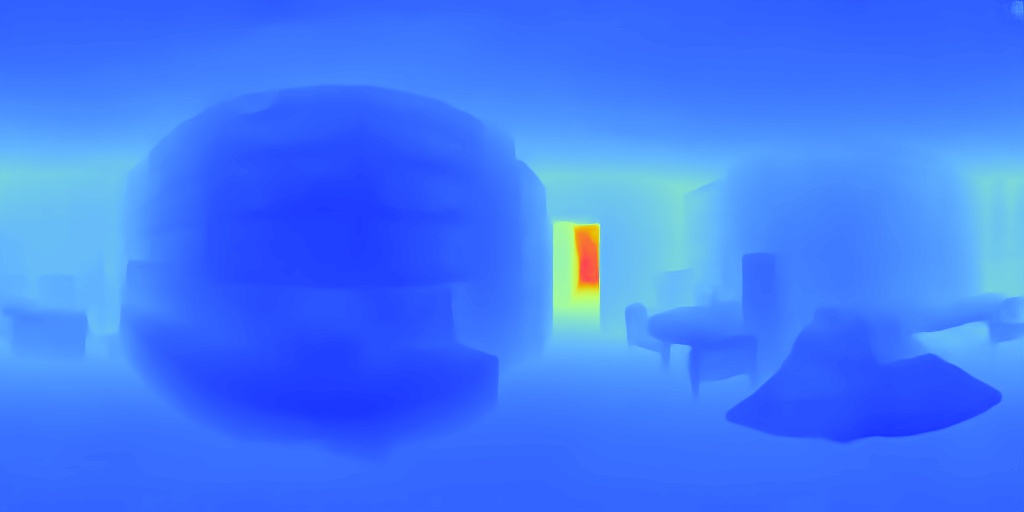} &
    \includegraphics[width=0.2\linewidth, trim={0px, 0px, 0px, 0px}, clip]{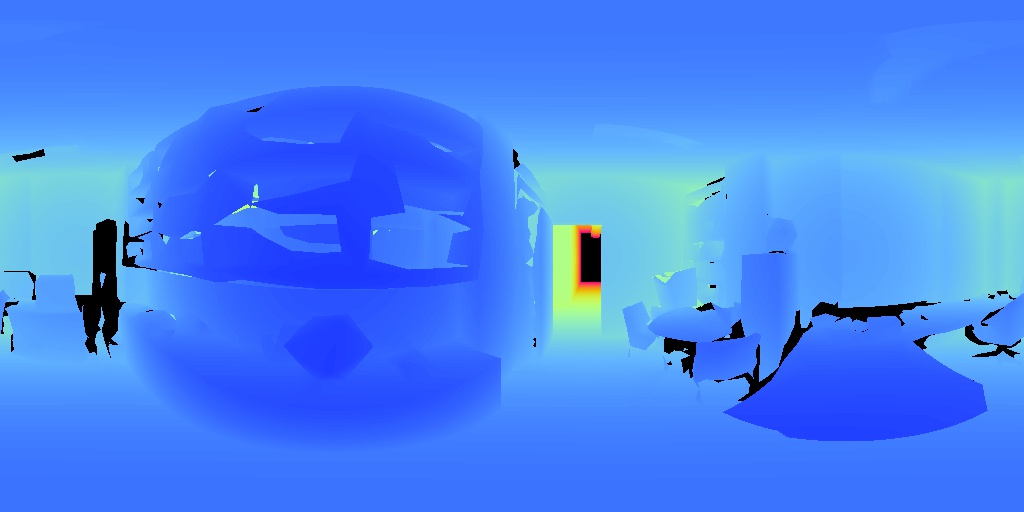}\\

      \small{Input Image} & \small{BiFuse~\cite{9157424}} & \small{SliceNet~\cite{Pintore_2021_CVPR}} &  \small{Ours} & \small{Ground Truth}
  \end{tabular}
  \end{center}
  
  \caption{Qualitative results on 360D (rows 1-3), Matterport3D (rows 4, 5) and Stanford2D3D (rows 6, 7). Invalid pixels are indicated with the black area in the ground truth.}
  \label{fig:3D60 result}
\end{figure*}

\begin{figure}[tbp]
  \begin{center}
  \renewcommand\tabcolsep{1.0pt}
  \begin{tabular}{cccc}
    \includegraphics[width=0.24\linewidth, trim={0px, 0px, 0px, 0px}, clip]{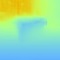}  &
    \includegraphics[width=0.24\linewidth, trim={0px, 0px, 0px, 0px}, clip]{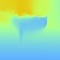}&
    \includegraphics[width=0.24\linewidth, trim={0px, 0px, 0px, 0px}, clip]{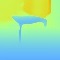}    &
    \includegraphics[width=0.24\linewidth, trim={0px, 0px, 0px, 0px}, clip]{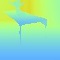}     \\
    
    \includegraphics[width=0.24\linewidth, trim={0px, 0px, 0px, 0px}, clip]{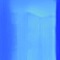}  &
    \includegraphics[width=0.24\linewidth, trim={0px, 0px, 0px, 0px}, clip]{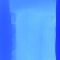}&
    \includegraphics[width=0.24\linewidth, trim={0px, 0px, 0px, 0px}, clip]{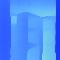}    &
    \includegraphics[width=0.24\linewidth, trim={0px, 0px, 0px, 0px}, clip]{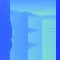}     \\
    
    \includegraphics[width=0.24\linewidth, trim={0px, 0px, 0px, 0px}, clip]{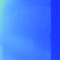}  &
    \includegraphics[width=0.24\linewidth, trim={0px, 0px, 0px, 0px}, clip]{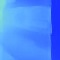}&
    \includegraphics[width=0.24\linewidth, trim={0px, 0px, 0px, 0px}, clip]{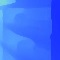}    &
    \includegraphics[width=0.24\linewidth, trim={0px, 0px, 0px, 0px}, clip]{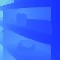}     \\

    \includegraphics[width=0.24\linewidth, trim={0px, 0px, 0px, 0px}, clip]{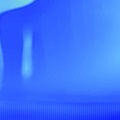}  &
    \includegraphics[width=0.24\linewidth, trim={0px, 0px, 0px, 0px}, clip]{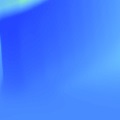} &
    \includegraphics[width=0.24\linewidth, trim={0px, 0px, 0px, 0px}, clip]{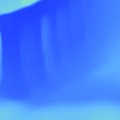} &
    \includegraphics[width=0.24\linewidth, trim={0px, 0px, 0px, 0px}, clip]{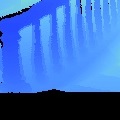}\\
    
    \includegraphics[width=0.24\linewidth, trim={0px, 0px, 0px, 0px}, clip]{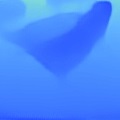}  &
    \includegraphics[width=0.24\linewidth, trim={0px, 0px, 0px, 0px}, clip]{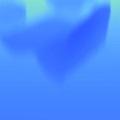} &
    \includegraphics[width=0.24\linewidth, trim={0px, 0px, 0px, 0px}, clip]{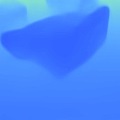} &
    \includegraphics[width=0.24\linewidth, trim={0px, 0px, 0px, 0px}, clip]{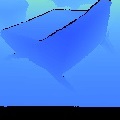}\\

    \includegraphics[width=0.24\linewidth, trim={0px, 0px, 0px, 0px}, clip]{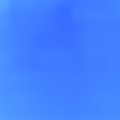}  &
    \includegraphics[width=0.24\linewidth, trim={0px, 0px, 0px, 0px}, clip]{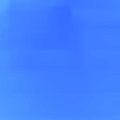} &
    \includegraphics[width=0.24\linewidth, trim={0px, 0px, 0px, 0px}, clip]{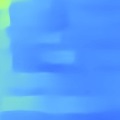} &
    \includegraphics[width=0.24\linewidth, trim={0px, 0px, 0px, 0px}, clip]{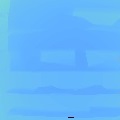}\\
    
    \includegraphics[width=0.24\linewidth, trim={0px, 0px, 0px, 0px}, clip]{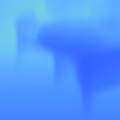}  &
    \includegraphics[width=0.24\linewidth, trim={0px, 0px, 0px, 0px}, clip]{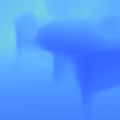} &
    \includegraphics[width=0.24\linewidth, trim={0px, 0px, 0px, 0px}, clip]{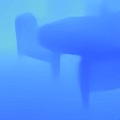} &
    \includegraphics[width=0.24\linewidth, trim={0px, 0px, 0px, 0px}, clip]{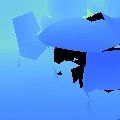}\\

      \small{BiFuse~\cite{9157424}} & \small{SliceNet~\cite{Pintore_2021_CVPR}} &  \small{Ours} & \small{Ground Truth}
  \end{tabular}
  \end{center}
  
  \caption{Qualitative comparison of the close-ups (marked by red boxes) in Fig.~\ref{fig:3D60 result}.}
  \label{fig:zoom_in_result}
\end{figure}

\subsection{Datasets and Evaluation Metrics}
We conduct our experiments mainly on three benchmark datasets: 360D~\cite{3D60}, Matterport3D~\cite{Matterport3D} and Stanford2D3D~\cite{Stanford3D}. 
Matterport3D and Stanford2D3D are real-world datasets collected by Matterport’s Pro 3D Camera. Matterport3D contains 10,800 panoramas and their corresponding depth maps. Stanford2D3D contains 1,413 panoramas collected from six large scale indoor areas of three kinds of buildings in the real world. As ToF sensors usually cause noise or missing value in certain areas, the ground-truth depth maps in Matterport3D and Stanford2D3D are incomplete and even inaccurate in some areas. Following many recent approaches \cite{9157424,Pintore_2021_CVPR}, we resize the resolution of images and depth maps in Matterport3D and Stanford2D3D into $512\times 1024$. We use official splits which take some rooms for training and the others for testing for both datasets. 360D comprises multi-modal stereo renders of scenes from realistic and synthetic large-scale 3D datasets including Matterport3D~\cite{Matterport3D}, Stanford2D3D~\cite{Stanford3D} and SunCG~\cite{SunCG}. 
In total, 360D contains 35,977 panoramas, where 33,879 of them are used for training, 800 used for validation and the rest is for testing. We use the split from Zioulis \emph{et al.}~\cite{3D60} and the resolution is resized to $512\times 256$.

To evaluate the performance, we use standard metrics (listed in Table \ref{tab:performance_comparison}) including mean absolute error (MAE), root mean square error (RMSE) and the root mean square error in log space (RMSElog). 
In addition, we calculate the percentages of pixels where the ratio between the estimated depth and ground truth depth is smaller than the thresholds $\delta$ to evaluate the accuracy.

\subsection{Comparisons}

For fair and reproducible experiments, we quantitatively and qualitatively compare our method with three baselines in this field: FCRN~\cite{Laina_2016_3DV}, OmniDepth~\cite{3D60} and BiFuse~\cite{9157424}. We also make comparisons with a state-of-the-art method SliceNet~\cite{Pintore_2021_CVPR} that also tries to handle rich global contexts in omnidirectional images. 

\textbf{Quantitative Comparisons }
We first show quantitative comparisons on three benchmark datasets in Table \ref{tab:performance_comparison}. 
The comparisons clearly show that our method improves state-of-the-art performance of panoramic depth estimation for most numerical metrics, especially on 360D and Matterport3D. On Stanford2D3D, our model performs comparably with SliceNet since this dataset only have 1,413 panoramas in total. For such a small-scale dataset, the potential of CViT cannot be fully exploited, since transformers favor large-scale datasets \cite{vit}.

\begin{figure}[tbp]
  \begin{center}
  \renewcommand\tabcolsep{1.0pt}
  \begin{tabular}{ccc}
    \includegraphics[width=0.333\linewidth, trim={0px, 0px, 0px, 0px}, clip]{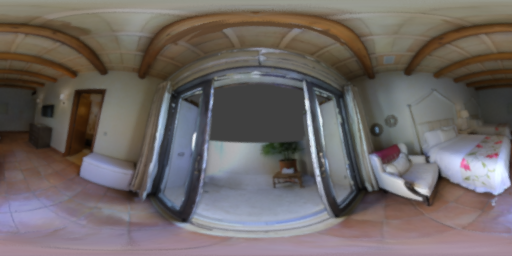}&
    \includegraphics[width=0.333\linewidth, trim={0px, 0px, 0px, 0px}, clip]{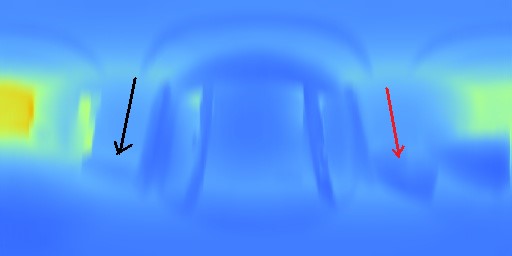}&
    \includegraphics[width=0.333\linewidth, trim={0px, 0px, 0px, 0px}, clip]{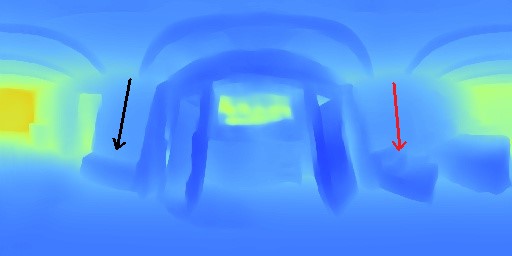} \\
    
    \includegraphics[width=0.333\linewidth, trim={0px, 0px, 0px, 0px}, clip]{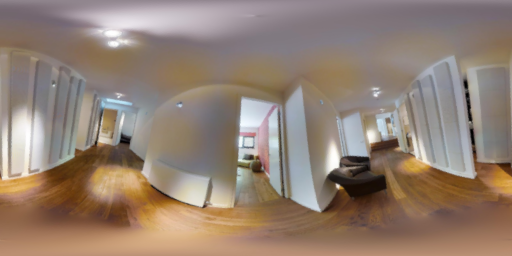}           &
    \includegraphics[width=0.333\linewidth, trim={0px, 0px, 0px, 0px}, clip]{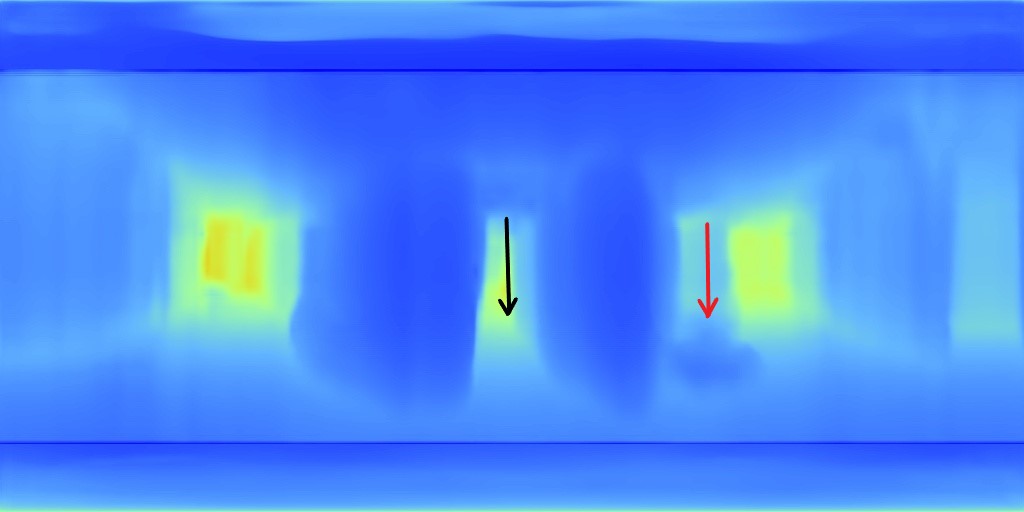}&
    \includegraphics[width=0.333\linewidth, trim={0px, 0px, 0px, 0px}, clip]{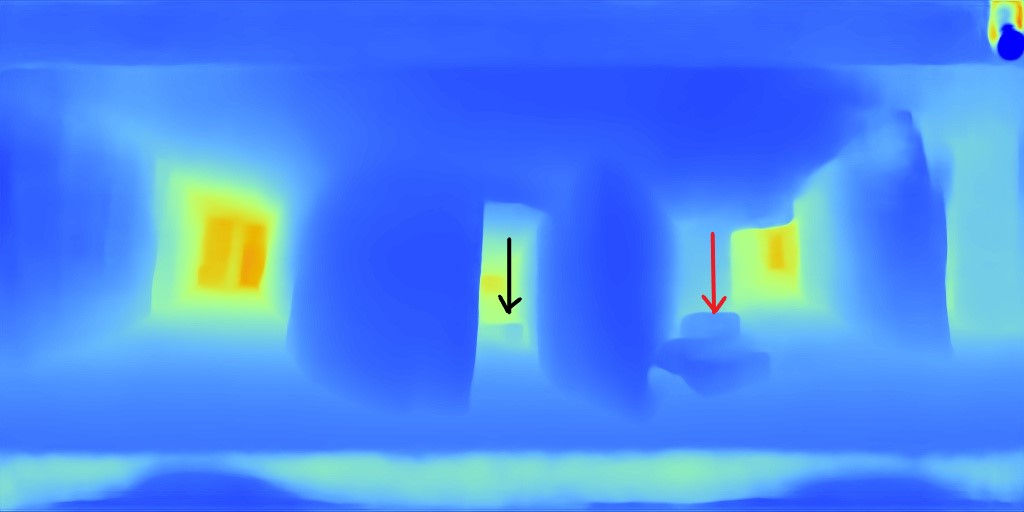}            \\
   \small{Input Image} & \small{SliceNet~\cite{Pintore_2021_CVPR}} &\small{Ours}
    
  \end{tabular}
  \end{center}
  
  \caption{Visually comparing our method with SliceNet \cite{Pintore_2021_CVPR} on non-gravity-aligned panoramas.}
  \label{fig:gravity_result}
\end{figure}

\textbf{Qualitative Comparisons }
In Figs.~\ref{fig:3D60 result} and \ref{fig:zoom_in_result}, we show visual comparisons against BiFuse and SliceNet.
Overall, with a fine-grained and globally coherent predictions, our GLPanoDepth performs favorably against state-of-the-art panoramic depth estimation methods.
Although cubemaps have already been used in some previous methods, \emph{e.g.}, BiFuse \cite{9157424}, their usages are fundamentally differently. Previous methods rely on cubemaps to handle the space-varying distortion of equirectangular projection. In contrast, we leverage cubemaps to extract global features by our specially-designed CViT. 
We observe that BiFuse tends to produce smooth predictions on test sets,
resulting in overly blur depth map. 
Based on the assumption that equirectangular projection is aligned to the gravity vector, SliceNet recovers panoramic depth maps through a convolutional long short-term memory (LSTM) network to retain the global information. As shown in the results, SliceNet enhances edges with the help of the global information. Instead of using LSTM, our GLPanoDepth leverages CViT branch to extract global features. MHSA in CViT is an inherently global operation, as every embedding token can attend to and thus influence other tokens globally. 
Compared with vertical spherical slices used in SliceNet, CViT takes cubemap representation as input, which suffers less from distortion and does not require the panorama to be aligned to the gravity vector. Therefore, our method works well even on non-gravity-aligned cases, as shown in Fig.~\ref{fig:gravity_result}. In this figure, two panoramas from the 360D dataset have been slightly warped with a certain degree, mimicking errors in the alignment. With such errors, the performance of SliceNet reduces obviously, while our method is robust to these cases, since no assumption is made by CViT.
Furthermore, our two-branch network can better learn global coherence in areas such as large homogeneous regions and relative depth arrangement across the image. 
Noted that our model can split the objects with a close depth plane, such as jars on the shelf. Our prediction produces cleaner and finer-grained delineations of object boundaries and in some cases less cluttered.

In Fig.~\ref{fig:teaser}, we show an example where our GLPanoDepth better preserves the geometric details of the scene, \emph{e.g.}, the chair and the bookshelf. This is because our method is aware of the global structures of the scene through CViT and can leverage global information to improve depth estimation.

\subsection{Ablation Study}
\begin{figure}[tbp]
  \begin{center}
  \renewcommand\tabcolsep{1.0pt}
  \begin{tabular}{cc}
    \includegraphics[width=0.47\linewidth, trim={0px, 0px, 0px, 0px}, clip]{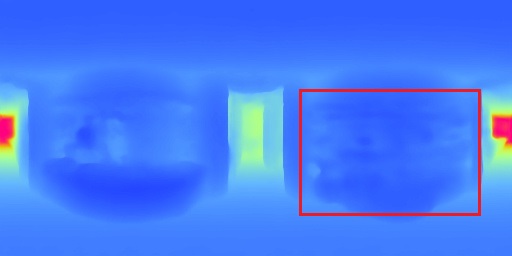}  &
    \includegraphics[width=0.47\linewidth, trim={0px, 0px, 0px, 0px}, clip]{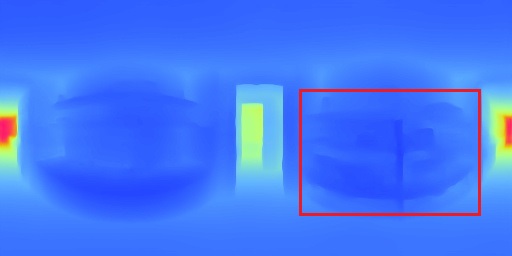}\\
    \small{Ours(CViT)} & \small{Ours(ViT+CNN)}\\
    \includegraphics[width=0.47\linewidth, trim={0px, 0px, 0px, 0px}, clip]{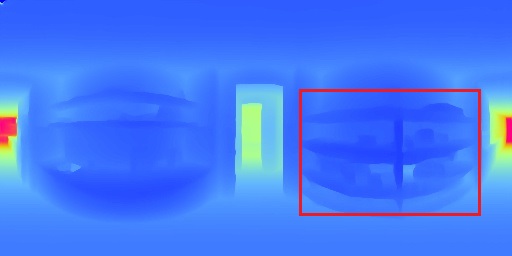}   &
    \includegraphics[width=0.47\linewidth, trim={0px, 0px, 0px, 0px}, clip]{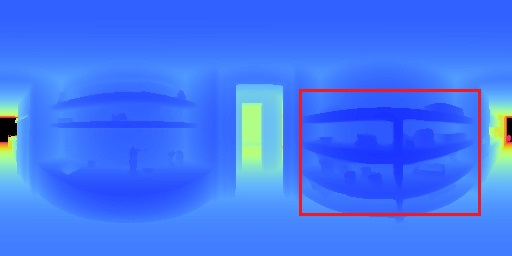} \\
    
     \small{Ours(Complete)} & \small{Ground Truth}\\     
    
\end{tabular}
\end{center}
  
  \caption{Visually comparing our complete model with two variants: Ours(CViT) removes the CNN branch and Ours(ViT+CNN) replaces CViT with ViT. The major differences are highlighted in red boxes.}
  \label{fig:ablation}
\end{figure}

\begin{figure}[tbp]
  \begin{center}
  \renewcommand\tabcolsep{1.0pt}
  \begin{tabular}{ccc}
  
    \includegraphics[width=0.33\linewidth, trim={0px, 0px, 0px, 0px}, clip]{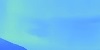}   &
    \includegraphics[width=0.33\linewidth, trim={0px, 0px, 0px, 0px}, clip]{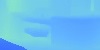}     &
    \includegraphics[width=0.33\linewidth, trim={0px, 0px, 0px, 0px}, clip]{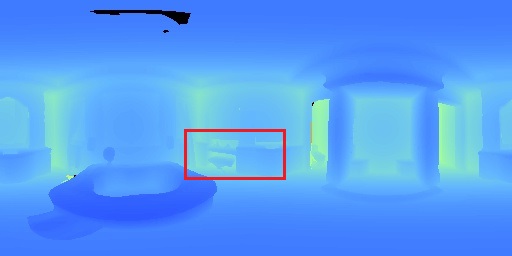}         \\
    
    \includegraphics[width=0.33\linewidth, trim={0px, 0px, 0px, 0px}, clip]{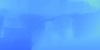}   &
    \includegraphics[width=0.33\linewidth, trim={0px, 0px, 0px, 0px}, clip]{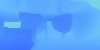}     &
    \includegraphics[width=0.33\linewidth, trim={0px, 0px, 0px, 0px}, clip]{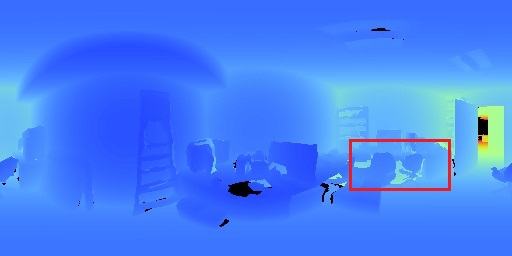}         \\
    \small{Concatenation} & \small{Gated fusion} & \small{Ground Truth}\\
    
\end{tabular}
\end{center}
  
  \caption{Validation of the effectiveness of the gated fusion module, as compared with simple concatenation.}
  \label{fig:ablation_fusion}
\end{figure}
We conduct various ablation experiments to investigate the effectiveness of our design choices, in particular, the two-branch architecture and the gated fusion module.

\textbf{Effectiveness of our two-branch architecture }
To validate the effectiveness of our two-branch architecture, we design two variants of our method: one removes the CNN branch from our GLPanoDepth and the other replaces CViT with the traditional ViT operated on panoramas directly.
As a single CNN branch cannot fully exploit the global information due to its limited receptive fields, GLPanoDepth employs an additional CViT branch in the whole framework. However, without the CNN branch, our method will also achieve sub-optimal results since a single CViT branch fails to preserve important and strong local features. In Ours(CViT), extracted features $\mathcal{F_G}$ from consecutive stages are combined using a RefineNet-based feature fusion block~\cite{RefineNet} and progressively upsampled by a factor of two to generate a fine-grained prediction. As illustrated in the up-left image of Fig.~\ref{fig:ablation}, many small structures are missing in the depth map predicted by Ours(CViT). The numerical metrics in Table~\ref{tab:performance_comparison} further indicate that a single CViT branch is hard to converge to the appropriate point with only extracted global features. Combining features from CViT branch and CNN branch offers highly synergistic improvements.

To study the advantage of CViT, we replace it with ViT. since ViT directly reshapes omnidirectional images into patches, it will suffer from distortion. The quantitative results in Table~\ref{tab:performance_comparison} show that using cubemaps reinforces the effect of transformer to extract global contents from spherical signals.

\textbf{Effectiveness of our gated fusion module }
To explore the effect of the gated fusion module, we remove it from our GLPanoDepth and use simple concatenation to combine features from different branches. Quantitative results in Table~\ref{tab:performance_comparison} clearly show that using concatenation as a fusion module makes our model converge to an inferior solution. As shown in Fig.~\ref{fig:ablation_fusion}, our gated fusion module yields sharper edges on the depth map, as well as clearer foreground-background separation, as compared with concatenation.
In addition, we observe that the model using our gated fusion module converges faster than that using concatenation, which further verifies the efficacy of our gated fusion module.

\begin{figure}[tbp]
  \begin{center}
  \renewcommand\tabcolsep{1.0pt}
  \begin{tabular}{ccc}

    \includegraphics[width=0.33\linewidth, trim={0px, 0px, 0px, 0px}, clip]{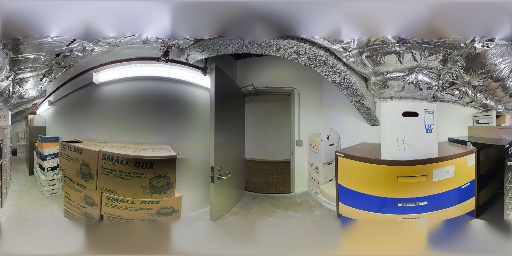}   &
    \includegraphics[width=0.33\linewidth, trim={0px, 0px, 0px, 0px}, clip]{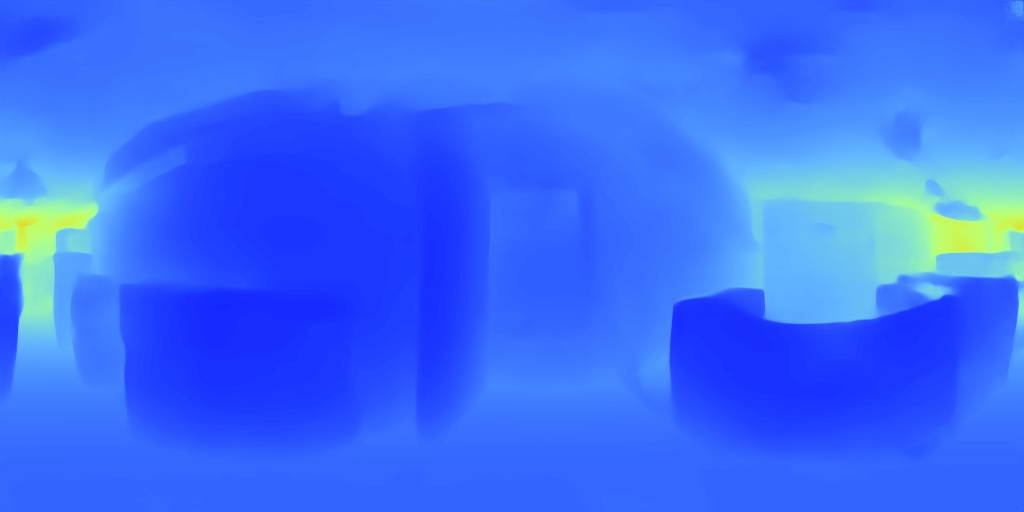}     &
    \includegraphics[width=0.33\linewidth, trim={0px, 0px, 0px, 0px}, clip]{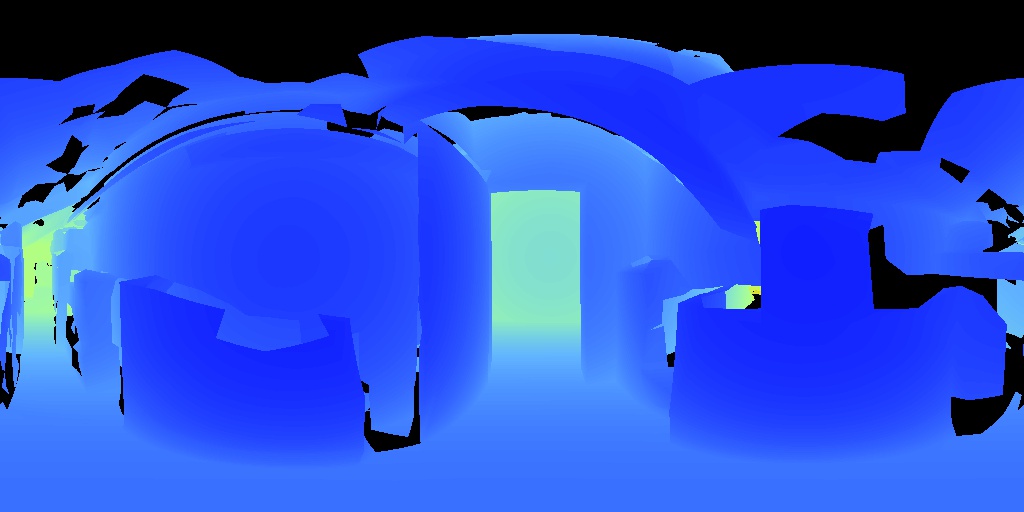}         \\
    \includegraphics[width=0.33\linewidth, trim={0px, 0px, 0px, 0px}, clip]{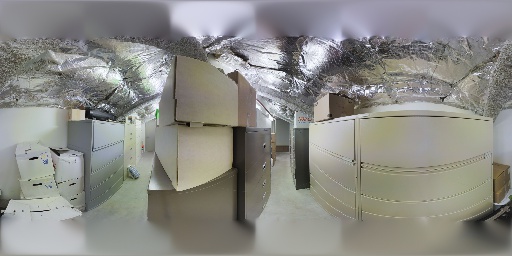}   &
    \includegraphics[width=0.33\linewidth, trim={0px, 0px, 0px, 0px}, clip]{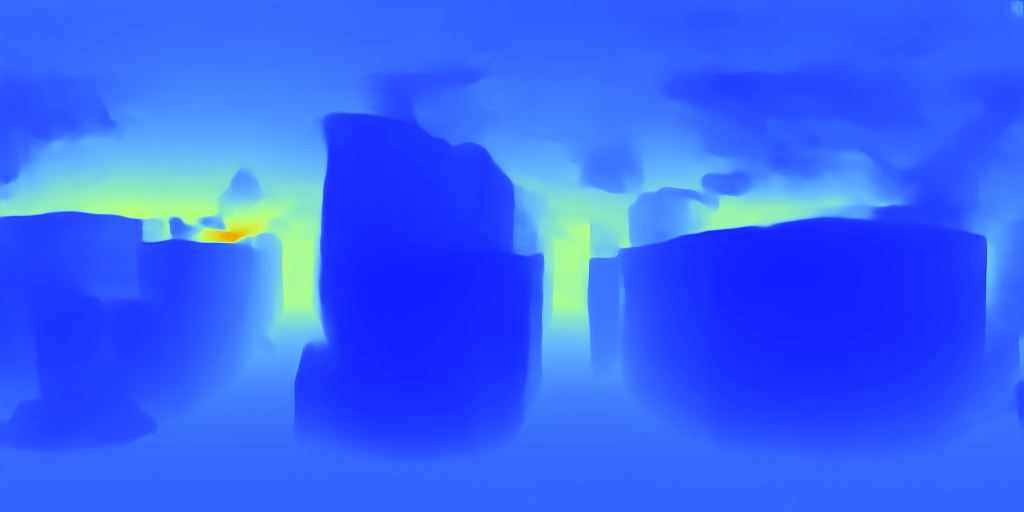}     &
    \includegraphics[width=0.33\linewidth, trim={0px, 0px, 0px, 0px}, clip]{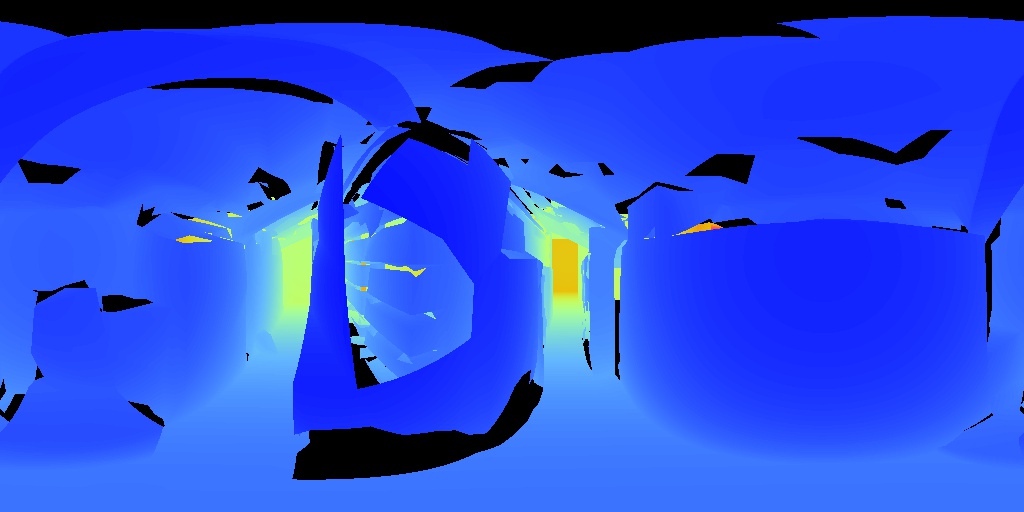}         \\
    \small{Input Image} & \small{Ours} & \small{Ground Truth}\\
    
\end{tabular}
\end{center}
  
  \caption{Two failure cases from the Stanford2D3D dataset.\label{fig:failure}}
\end{figure}

\subsection{Limitations}
We have shown how the proposed CViT architecture is beneficial for extracting rich global information from omnidirectional images. However, like other transformers \cite{vit}, CViT requires a large-scale dataset for training. In other words, if the dataset is small in size, the accuracy of prediction will decrease. As we have mentioned previously, our method with a single CViT branch fails to converge on the Stanford2D3D dataset, since this dataset only has 1,413 panoramas in total, most of which contain large areas of missing depth value. Two failure cases from the Stanford2D3D dataset are shown in Fig.~\ref{fig:failure}. To maximize the effectiveness of CViT and to improve the generalization capability of the trained model, a large-scale dataset containing a large amount of diverse training data is needed.

\section{Conclusions}
To conclude, we have proposed GLPanoDepth, a new method for end-to-end panoramic depth estimation. The key idea of GLPanoDepth is to extract global and local features respectively by two separate branches: one is based on cubemap vision transformers and the other is based on traditional convolutional layers. Different features are progressive combined by a gated fusion module.
We validate the benefits of our proposed method on multiple datasets. Qualitative and quantitative comparisons against previous methods show superiority of our method.
Ablation studies further exemplify that the specially-designed cubemap vision transformers and gated fusion module are able to better capture rich and distortion-free global features from spherical signals compared with other methods.

{\small
\bibliographystyle{ieee_fullname}
\bibliography{egbib}
}

\end{document}